\documentclass[10pt,twocolumn,letterpaper]{article}

\usepackage[pagenumbers]{cvpr} 

\definecolor{cvprblue}{rgb}{0.21,0.49,0.74}
\usepackage[pagebackref,breaklinks,colorlinks,allcolors=cvprblue]{hyperref}
\usepackage{hyperref}
\usepackage{url}
\usepackage{multirow}
\usepackage{array}
\usepackage{booktabs}
\usepackage{xcolor} 
\usepackage{colortbl}
\usepackage{graphicx} 
\usepackage{subcaption}
\usepackage{pifont}
\usepackage{booktabs}
\usepackage{colortbl}

\def\paperID{18593} 
\def\confName{CVPR}
\def\confYear{2026}

\title{TimeScope: Towards Task-Oriented Temporal Grounding In Long Videos} 

\author{Xiangrui Liu$^{1,2\dagger}$ \and 
        Minghao Qin$^{1\dagger}$ \and 
        Yan Shu$^{3\dagger}$ \and 
        Zhengyang Liang$^{4}$ \and 
        Yang Tian$^{2}$ \and 
        Chen Jason Zhang$^{1}$ \and 
        Bo Zhao$^{2}$ \and 
        Zheng Liu$^{1}$ \and \\
        $^1$Beijing Academy of Artificial Intelligence
        $^2$School of AI, Shanghai Jiao Tong University\\
        $^3$University of Trento
        $^4$Singapore Management University \\
        \small $\dagger$Equal contribution.
}

\begin{document}
\maketitle
\begin{abstract}
Identifying key temporal intervals within long videos, known as temporal grounding (TG), is important to video understanding and reasoning tasks. In this paper, we introduce a new form of the temporal grounding problem, \textbf{Task-oriented Temporal Grounding} (\textbf{ToTG}), which is driven by the requirements of downstream tasks rather than explicit time-interval descriptions. For example, a ToTG input may be “explain why the man in the video is sent to the hospital,” whereas traditional TG would take an explicit temporal description such as “the moments when the man is tripped by a stone and falls to the ground.” This new ToTG formulation presents significant challenges for existing TG methods, as it requires jointly performing deep task comprehension and fine-grained temporal localization within long videos. To address these challenges, we conduct a systematic set of studies. First, we construct \textbf{a new benchmark ToTG-Bench}, which comprehensively evaluates ToTG performance across diverse settings. Second, we introduce \textbf{a new temporal-ground method TimeScope}, which performs coarse-to-fine localization through a progressive reasoning process. Leveraging extensive supervised fine-tuning with carefully curated chain-of-thought (CoT) data from a variety of scenarios, TimeScope generalizes effectively across tasks and domains. Our evaluation demonstrates \textbf{TimeScope’s empirical advantages} over existing baselines from three perspectives: (1) substantial improvements in grounding precision, (2) significant benefits to downstream tasks, and (3) strong generalizability across different scenarios. All models, datasets, and source code will be fully open-sourced to support future research in this area.

\end{abstract}
\begin{figure*}[]
    \centering
    \includegraphics[width=\linewidth]{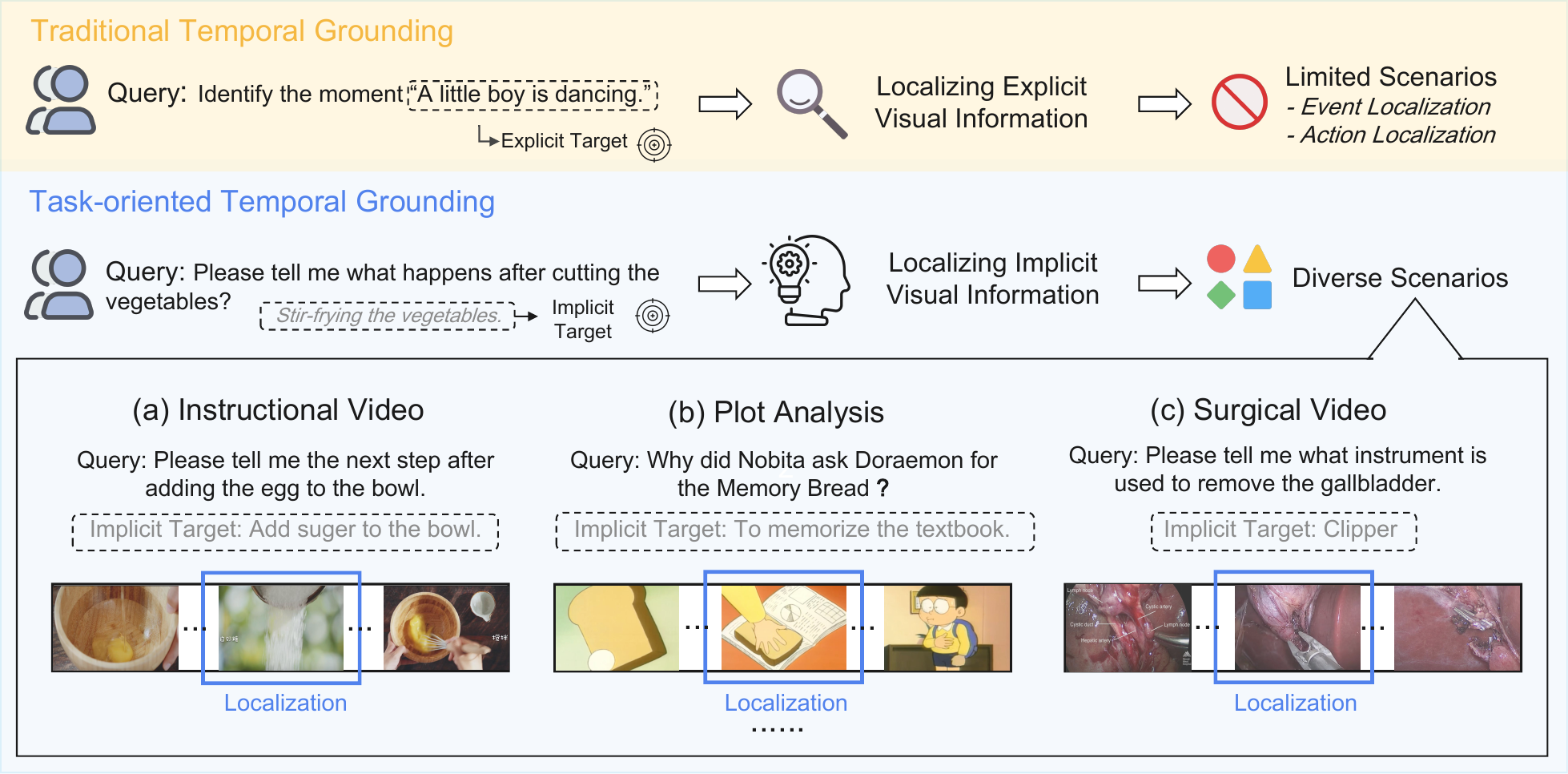}
    \caption{In traditional temporal grounding, the target is explicit and can be located via simple semantic matching, whereas task-oriented temporal grounding requires identifying an implicit target essential for completing the task.}
    \label{fig:definition}
\end{figure*} 


\section{Introduction}
Multimodal large language models (MLLMs) have become increasingly prominent in tackling long-video understanding (LVU) problems. However, they still struggle with complex tasks which call for fine-grained details, especially those sparsely distributed across long videos. One promising strategy to mitigate this problem is to present MLLMs only with crucial temporal intervals which contains relevant information to their LVU tasks. However, existing temporal grounding (TG) methods are primarily designed for tasks with explicit descriptions of time-intervals, such as “\textit{the moments when the man is tripped by a stone and falls to the ground}”, instead of directly handling native task requirements, e.g., “\textit{explain why the man in the video is sent to the hospital}”. This gap prevents the direct utilization of existing TG methods for LVU tasks, resulting in a severe limitation for real-world applications.

To formalize the above challenge, we define a new problem: {Task-oriented Temporal Grounding} (\textbf{ToTG}), where a model needs to localize crucial temporal intervals that are relevant to a specific downstream task based on its native requirement. For example, as illustrated in Figure~\ref{fig:definition}, given the query “\textit{tell me what happens after cutting the vegetables},” the model identifies the interval corresponding to “\textit{stir-frying the vegetables}.” Unlike traditional TG problems, ToTG is more closely aligned with practical applications, as it can be directly conducted to support real-world LVU tasks. The new ToTG problem introduces unprecedented technical challenges for existing methods on TG, as it requires both \textit{in-depth task comprehension} and \textit{fine-grained temporal localization} over long videos.  

In this paper, we present a systematic study of the ToTG problem. First, to address the absence of suitable resources for evaluating ToTG performance, we introduce a new benchmark, \textbf{ToTG-Bench}. ToTG-Bench incorporates 32 diverse video domains and 12 task categories, with video durations ranging from a few seconds to over an hour. Each instance is constructed through a human–machine collaborative annotation pipeline that ensures high-quality temporal localization. Together, these designs enable ToTG-Bench to provide a comprehensive evaluation of ToTG performance across a wide spectrum of task and video types.  

To further enhance the grounding model’s capability for ToTG, we develop a novel progressive-reasoning framework called \textbf{TimeScope}. As discussed, ToTG poses two major technical challenges: 1) in-depth understanding of the task, and 2) fine-grained temporal grounding within long videos. TimeScope addresses these challenges through two consecutive operations. First, the model focuses on comprehending the task and identifying which parts of the video are likely to contain the required information. To accomplish this, TimeScope performs chain-of-thought (CoT) reasoning based on the task description and a holistic abstraction of the video, producing a set of candidate time scopes that are likely to be relevant. Next, TimeScope further encodes the candidate scopes with more detailed visual information, and predicts the precise temporal intervals contained within them.
This progressive reasoning pipeline not only improves the grounding precision but also enables more efficient processing of long videos. 

To obtain a general temporal-grounding capability across diverse long-video tasks, TimeScope is trained through extensive supervised fine-tuning. To construct the training data, we prompt a teacher model to generate chain-of-thought (CoT) reasoning for a wide range of LVU tasks. We then filter the generated CoT trajectories, retaining only those whose reasoning leads to precise temporal grounding. This process yields \textbf{ToTG-Pile}, a diverse and high-quality dataset tailored for ToTG.

In our experiments, we evaluate TimeScope against traditional TG methods~\cite{bai2025qwen25vltechnicalreport, yang2025kwaikeyevl15technical} using not only the proposed ToTG-Bench, but also standard temporal-grounding benchmarks~\cite{caba2015activitynet, sigurdsson2016Charades, cheng2025vstarbenchmarkingvideollmsvideo} and long-video understanding benchmarks~\cite{wu2024longvideobench, zhou2025mlvu}. Our evaluation demonstrates the effectiveness of TimeScope from three perspectives. First, it substantially improves temporal-grounding precision over existing TG baselines. Second, it exhibits strong generalizability, achieving consistent performance gains across heterogeneous benchmarks. Third, it significantly enhances the performance of downstream LVU tasks when used as a pre-localization module.
Extended ablation studies further highlight the contribution of each component, reflecting the validity of our technical design. All resources, including the model, benchmark, dataset, and source code, will be publicly released to facilitate future research.

The contributions of this paper are summarized as follows. 1) We formulate the \textbf{ToTG problem}, which formally defines the localization of task-relevant information in long videos. 2) We introduce \textbf{ToTG-Bench}, which enables unified and comprehensive evaluation of ToTG performance across diverse tasks and video domains. 3) We propose \textbf{TimeScope}, a progressive reasoning framework for ToTG, and construct \textbf{ToTG-Pile}, a high-quality supervised fine-tuning dataset that enhances TimeScope’s capability. 4) We conduct extensive experiments, demonstrating TimeScope’s \textbf{empirical effectiveness} in ToTG precision, generalization across benchmarks, and benefits to downstream LVU tasks.

\begin{table*}[t]
\centering
\begin{tabular}{l|ccccc}
\toprule
{Benchmark} & {Video Num.} & {Avg. Duration} & {Duration Range} & {Video Domain}  & {Query Type} \\
\midrule
\rowcolor{gray!20} \multicolumn{6}{c}{{Traditional Temporal Grounding Benchmark}} \\
Charades-STA  & 1,331 & 29.9 s & 7.2 s - 1.2 min & Daily Activities & Explicit Description \\
ActivityNet  & 4,885 & 122.0 s & 2.3 s - 12.4 s & Daily Activities & Explicit Description \\
V-STaR & 732 & 1.8 min & 15.0 s - 59.2 min & 9 Domains & Explicit Description \\
\rowcolor{gray!20} \multicolumn{6}{c}{{Clue-grounded QA Benchmark }} \\
CG-Bench & 1,219 & 28.7 min & 9.1 min - 1.8 hr & 14 Domains & Perception Task \\
Next-GQA (test) & 990 & 38.7 s & 10.0 s - 2.5 min & Daily Activities & Perception Task \\
\rowcolor{gray!20} \multicolumn{6}{c}{{Task-oriented Temporal Grounding Benchmark }} \\
ToTG-Bench & 337 & 13.5 min & 30 s - 1.2 hr & 35 Domains  & Perception/Reasoning Task \\
\bottomrule
\end{tabular}
\vspace{-8px} 
\caption{Comparison of ToTG-Bench with previous temporal grounding and clue-grounded QA benchmarks. ToTG-Bench demonstrates superior diversity and comprehensiveness in video characteristics and query types.}
\label{tab:totgbench}
\end{table*}

\section{Related work}
\subsection{Long Video Understanding}
\label{sec:problem}
The field of long video understanding (LVU) has developed rapidly in recent years, with many powerful MLLMs emerging, such as VideoChatFlash~\cite{li2025videochatflashhierarchicalcompressionlongcontext}, Video-XL-2~\cite{qin2025videoxl2longvideounderstandingtaskaware}, Eagle2.5~\cite{chen2025eagle25boostinglongcontext}, and InternVL3~\cite{zhu2025internvl3exploringadvancedtraining}. These models demonstrate strong general video understanding capabilities and serve as versatile backbones for various video tasks.  
However, precisely capturing fine-grained details within second-level intervals remains a major challenge for current LVU models. To address this, some works introduce additional modules to assist LVU models by identifying key frames~\cite{wang2025videoitg,huang2025frag,yu2024frame,qin2025task}. These modules are typically similarity-based and thus lack deeper semantic understanding of the video content, limiting their compatibility with diverse downstream tasks in long video scenarios.  
In contrast, we take a different approach. We post-train LVU MLLMs on our diverse and high-quality task-oriented grounding dataset, and further implement \textbf{TimeScope}, a novel framework designed for progressive task-oriented grounding. This enables the model to efficiently and accurately localize critical time intervals in long videos for a wide range of tasks.


\subsection{Video Temporal Grounding}
The traditional temporal grounding (TG) task requires models to localize a time interval in a video given a query that explicitly describes the target content. Early approaches are mainly dual-encoder-based, where video and language features are extracted using different pre-trained encoders (e.g., BERT~\cite{devlin2019bertpretrainingdeepbidirectional}, CLIP~\cite{radford2021learningtransferablevisualmodels}, SigLip~\cite{zhai2023sigmoidlosslanguageimage}), and then fused for time interval xprediction~\cite{CVPR2024SnAG,lei2021detecting,moon2023query,moon2023correlation,gordeev2024saliency,song2024moviechatquestionawaresparsememory,song2024moviechatdensetokensparse}. These models lack generalizability and can only be evaluated under few-shot setting across different benchmarks.
More recently, researchers have explored using MLLMs for more general temporal grounding~\cite{huang2024lita, ren2024timechat, zeng2025timesuite,guo2024trace,wang2025timer1posttraininglargevision}. For instance, TimeChat~\cite{ren2024timechat} introduces a time-aware frame encoder that binds visual tokens with their corresponding timestamps at the frame level for temporal grounding. Similarly, TimeSuite~\cite{zeng2025timesuite} proposes temporal-adaptive position encoding to strengthen temporal awareness in video representations. Trace~\cite{guo2024trace} designs a specialized encoder and head for timestamp input, while Time-R1~\cite{wang2025timer1posttraininglargevision} employs a reasoning-guided post-training framework with reinforcement learning and verifiable rewards to improve grounding accuracy. In addition to these specialized MLLMs, recent generic MLLMs (e.g., Qwen2.5-VL~\cite{bai2025qwen25vltechnicalreport}, Keye-VL-1.5~\cite{yang2025kwai}) have also demonstrated certain capabilities for temporal grounding.  
However, localizing small intervals in long videos is challenging due to their limited context window. UniTime~\cite{li2025universal} tackles this by adjusting the frame sampling rate for multi-stage grounding, but its sparse frame sampling can disrupt continuous event semantics, compromising precise temporal grounding.
In addition, previous approaches tend to fall short when it comes to more complex and practical grounding tasks.
Motivated by these limitations, we introduce the new problem of \textbf{task-oriented temporal grounding}, along with a benchmark, a dataset, and a dedicated framework to address it.

\begin{figure*}[t]
    \centering
    \includegraphics[width=\linewidth]{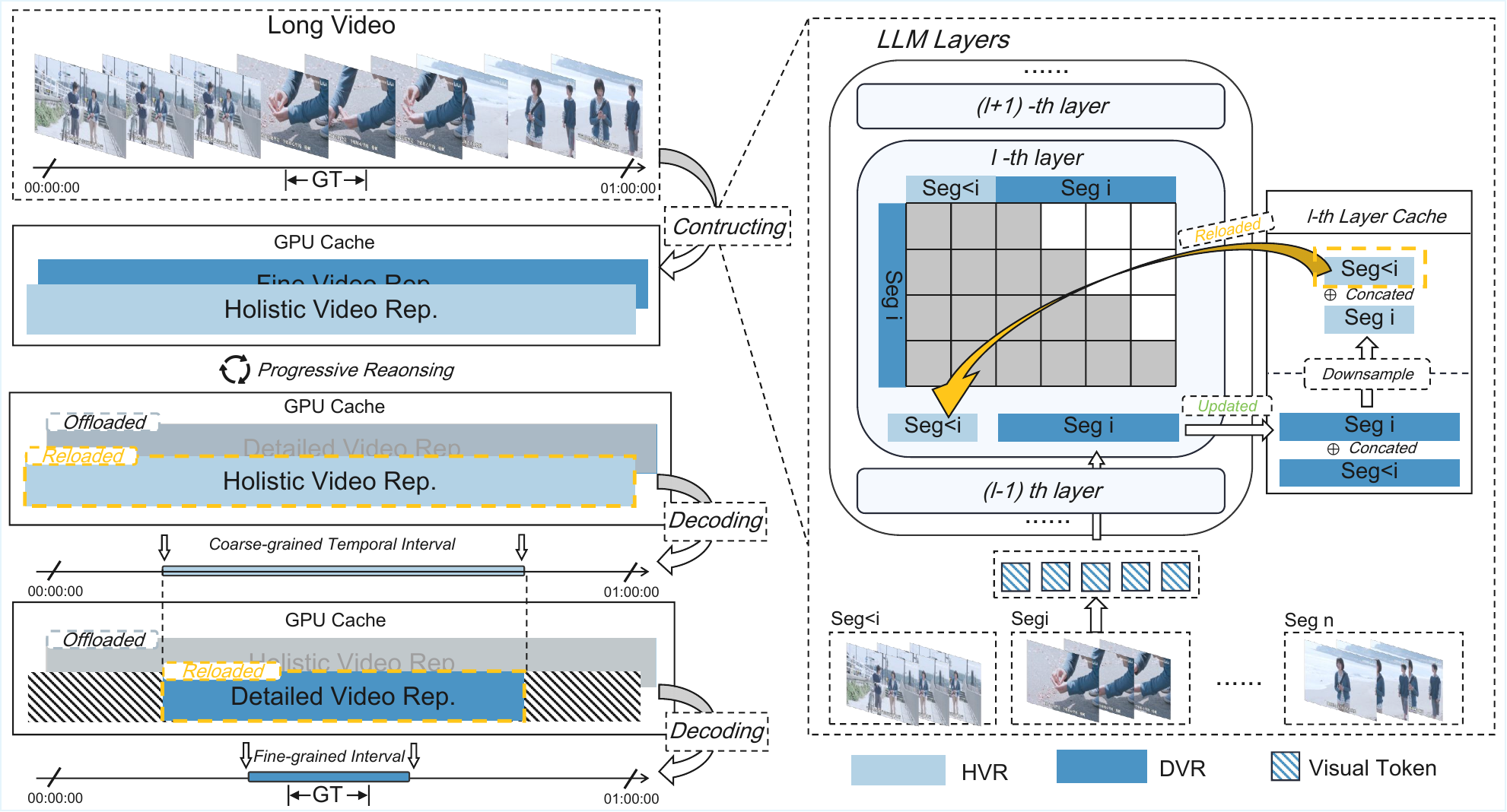}
    \caption{
    Overview of TimeScope. The input long video is processed to generate two representations: the Holistic Video Representation (HVR), which captures global context, and the Fine Video Representation (FVR), which retains detailed local information. TimeScope first performs coarse-grained reasoning using HVR to narrow the search space, and then refines the localization using FVR within the identified temporal interval to achieve precise task-oriented localization.}
    \label{fig:framework}
\end{figure*}

\section{Problem Definition}
\label{sec:problem}
We formally define the Task-oriented Temporal Grounding (ToTG) problem in this section. Formally, the input consists of a long video 
$V=\{f_t\}_{t=1}^T$ 
and a task-oriented natural language query $Q_\text{task}$.
ToTG assumes that the critical visual information required to answer the task is not explicitly described in $Q_{\text{task}}$. Therefore, the model must localize the temporal interval containing this task-critical but implicit evidence \emph{without relying on direct semantic matches to the query}.
Let the target interval be denoted as $[start^{\*}, end^{\*}]$. We define ToTG as:
\[
[start^{\*}, end^{\*}] = \Gamma(V, Q_\text{task}),
\]
where $\Gamma$ is a grounding function that selects the video segment containing the necessary task-related visual cues, even though such cues are not directly mentioned in the query.
This formulation represents a broad variety of real-world scenarios (such as instructional videos, plot reasoning, and surgical workflows) where the required visual evidence is implicit and must be located in the video rather than retrieved through explicit semantic matching.

\section{Method}
This section outlines the benchmark \textbf{ToTG-Bench}, the proposed framework \textbf{TimeScope}, and the specialized dataset \textbf{ToTG-Pile} in Sec.~\ref{sec:benchmark}, Sec.~\ref{sec:framework}, and Sec.~\ref{sec:dataset}, respectively.

\subsection{ToTG-Bench}
\label{sec:benchmark}
To enable a comprehensive evaluation of ToTG, we introduce \textbf{ToTG-Bench}.
We construct ToTG-Bench based on long-video QA datasets, since their most questions naturally align with the definition of ToTG—queries that implicitly specify task-required information within extended video contexts.
Moreover, such datasets cover a wide range of domains and task types, allowing us to build a benchmark that is both diverse and realistic.
Concretely, we collect question–answer pairs from four public long-video understanding datasets~\cite{zhou2025mlvu, fu2025video, wu2024longvideobench, cheng2025v}, using the questions as task instruction style queries $Q_\text{task}$ and deriving the corresponding ground-truth temporal intervals through a semi-automated annotation pipeline. First, Gemini-2.5-Pro is used to predict multiple temporal intervals as candidate segments. These candidates are then carefully verified, selected, and refined manually to produce the final grounded annotations.
As shown in Table~\ref{tab:totgbench}, ToTG-Bench exhibits substantially higher diversity and comprehensiveness compared with both traditional temporal grounding benchmarks (Charades-STA~\cite{sigurdsson2016Charades}, ActivityNet~\cite{caba2015activitynet}, and V-STaR~\cite{cheng2025v}) and clue-grounded QA benchmarks (CG-Bench~\cite{chen2024cg} and Next-GQA~\cite{xiao2024can}). 
It covers a wide range of video durations—from seconds to over one hour—with an average length of 13.5 minutes, and spans 35 realistic video domains (e.g. vlogs, news, documentaries, and sports). 
Moreover, ToTG-Bench incorporates $Q_{task}$ from both perception tasks (e.g., action or object recognition) and reasoning tasks (e.g., temporal or causal reasoning), covering a total of 12 distinct task types.
These characteristics make ToTG-Bench a diverse, comprehensive, and realistic foundation for benchmarking task-oriented temporal grounding models. More details about ToTG-Bench can be found Appendix.

\subsection{TimeScope}
\label{sec:framework}
We propose \textbf{TimeScope}, a progressive reasoning framework that reduces the search space in a coarse-to-fine manner: it first identifies a coarse temporal interval where the target is likely to occur, and then performs refined grounding within this narrowed region.

\textbf{Holistic \& Detailed Representation.}
Dense frame sampling is crucial for accurate grounding in long videos, as it preserves fine-grained temporal continuity and reduces the risk of missing brief task-critical moments. However, existing methods such as UniTime~\cite{li2025universal} are limited to sparse sampling due to the prohibitive memory cost of processing long sequences.
To reconcile dense sampling with memory constraints, TimeScope decouples global context from local details. Specifically, we define a \emph{Holistic Video Representation (HVR)} to abstract long-range context with minimal memory overhead, and a \emph{Detailed Video Representation (DVR)} to retain high-resolution visual cues for precise localization.

To avoid the memory overflow of processing all frames simultaneously, TimeScope constructs both representations in a streaming manner.
As illustrated in Fig.~\ref{fig:framework} (right), the densely sampled video is divided into temporal segments ${S_0, S_1,\dots,S_n}$. Each segment $S_i$ is sequentially processed by the MLLM to produce both representations. For each layer $l$, the key–value states computed from $S_i$ form its detailed representation $DVR_i^l$. These KVs are then temporally downsampled to obtain the holistic counterpart $HVR_i^l$, which summarizes segment-level semantics at significantly reduced resolution.

After that, all previously generated holistic caches ${HVR_j^l\mid j<i}$ are concatenated to form a lightweight historical memory. The current detailed representation $DVR_i^l$ attends to this holistic memory through cross-attention, enabling efficient integration of long-range temporal information without incurring full computation on the entire video. The resulting updated states are forwarded to obtain $DVR_i^{l+1}$.
After all segments are processed, the per-segment representations ${DVR_i}$ and ${HVR_i}$ are concatenated along the temporal dimension to form the final detailed- and holistic-level summaries of the entire video, which are stored for subsequent progressive reasoning.

\begin{table*}[h]
\centering
\fontsize{8}{10}\selectfont 
\setlength{\tabcolsep}{1pt} 
\begin{tabular}{l|cccccc}
\midrule
\textbf{Method} & \textbf{S(R1@0.3)} & \textbf{S(R1@0.5)} & \textbf{M(R1@0.3)} & \textbf{M(R1@0.5)} & \textbf{L(R1@0.3)} & \textbf{L(R1@0.5)} \\
\midrule
\rowcolor{gray!15}\multicolumn{7}{c}{\textbf{Temporal Grounding Models }} \\
TimeR1-7B~\cite{wang2025timer1posttraininglargevision} & 20.1 & 16.0 & 6.8 & 4.1 & 15.8 & 11.0 \\
VideoChatR1-7B~\cite{li2025videochatr1enhancingspatiotemporalperception} & 17.3 & 14.0 & 6.6 & 5.6 & 12.1 & 10.9 \\
Temporal-RLT-7B~\cite{li2025reinforcementlearningtuningvideollms} & 54.0 & 35.3 & 17.9 & 12.3 & 14.1 & 11.5 \\
UniTime~\cite{li2025universalvideotemporalgrounding} & 52.4 & 42.7 & 29.6 & 23.2 & 24.2 & 22.6 \\
\rowcolor{gray!15}\multicolumn{7}{c}{\textbf{Video Understanding Models }} \\
Qwen-2.5VL-7B~\cite{bai2025qwen25vltechnicalreport} & 48.2 & 40.6 & 12.2 & 10.3 & 14.6 & 12.3 \\
Keye-VL-1.5-8B~\cite{yang2025kwaikeyevl15technical} & 36.0 & 34.7 & 25.4 & 25.4 & 32.7 & 28.6 \\
\rowcolor{gray!20}
\textbf{TimeScope-7B} & \textbf{52.3} & \textbf{46.3} & \textbf{45.2} & \textbf{42.3} & \textbf{47.3} & \textbf{37.8}  \\
\hline
\end{tabular}
\vspace{-8px} 
\caption{Performance comparison on ToTG-bench. ``S" refer to "Short", ``M" refer to ``Medium", ``L" refer to ``Long". ``has option” represents incorporating the options into the prompt.}
\vspace{-8px} 
\label{tab:lvtgbench}
\end{table*}




\begin{table*}[h]
\centering
\fontsize{8}{10}\selectfont
\setlength{\tabcolsep}{3pt}
\begin{tabular}{l|cccc|ccccccccccc}
\midrule
\textbf{Model} & \textbf{OR} & \textbf{SR} & \textbf{TR} & \textbf{AR} & \textbf{SP} & \textbf{AP} & \textbf{TP} & \textbf{EG} & \textbf{OP} & \textbf{OB} & \textbf{AT} & \textbf{CP} & \textbf{TU} \\
\midrule
TimeR1-7B & 13.8 & 17.2 & 11.8 & 6.50 & 2.90 & 14.8 & 8.80 & 8.20 & 7.70 & 17.4 & 7.70 & 13.3 & 12.5 \\
Temporal-RLT-7B & 22.4 & 20.7 & 23.5 & 8.70 & \textbf{42.9} & 48.1 & 23.5 & 6.40 & 26.9 & 23.9 & 10.8 & 40.0 & 6.20 \\
Qwen-2.5VL-7B & 24.1 & 34.5 & 23.5 & 19.6 & 1.20 & 40.7 & 17.6 & 10.0 & 30.8 & 28.3 & 18.7 & 33.3 & 9.30 \\
Keye-VL-1.5-8B & 19.0 & 27.6 & 35.3 & 41.3 & 10.7 & 22.2 & \textbf{55.9} & 10.0 & 26.9 & 30.4 & 36.9 & 26.7 & \textbf{56.2} \\
\rowcolor{gray!20}
\textbf{TimeScope-7B} & \textbf{42.9} & \textbf{48.3} & \textbf{47.1} & \textbf{47.8} & 36.7 & \textbf{63.0} & 35.3 & \textbf{30.0} & \textbf{46.2} & \textbf{42.9} & \textbf{36.9} & \textbf{40.0} & 35.8 \\

\midrule
\end{tabular}
\vspace{-8px}
\caption{Performance comparison on different task types in the ToTG-bench. All values are rounded to three decimal places. The metric is IoU@0.5.
OR: Object Reasoning, SR: Spatial Reasoning, TR: Temporal Reasoning, AR: Action Reasoning, SP: Spatial Perception, AP: Attribute Perception, TP: Temporal Perception, EG: Ego, OP: OCR Problems, AT: Action Recognition, CP: Counting Problem, TU: Tutorial.}
\label{tab:main_result_1}
\end{table*}

\begin{table*}[t]
\centering
\fontsize{8}{10}\selectfont 
\begin{tabular}{l|ccc|ccc}
\hline
\multirow{2}{*}{\textbf{Method}} & \multicolumn{3}{c|}{\textbf{Charades-STA}} & \multicolumn{3}{c}{\textbf{ActivityNet}} \\
 & R1@0.5 & R1@0.7 & mIoU & R1@0.5 & R1@0.7 & mIoU \\
\midrule
\rowcolor{gray!15}\multicolumn{7}{c}{\textbf{Open-source VLP Method}} \\ 

2D-TAN~\cite{zhang20192DTAN}  & 45.8 & 27.9 & -- & 60.4 & 43.4 & --  \\
UniVTG~\cite{lin2023univtg}  & 60.2 & 38.6 & -- & 56.1 & 43.4 & --  \\
SSRN~\cite{ssrn}  & 65.5 & 42.6 & -- & -- & 54.5 & --  \\
SnAG~\cite{CVPR2024SnAG}  & 64.6 & 46.2 & -- & -- & 48.6 & -- \\
EaTR~\cite{jang2023knowing}  & 68.4 & 44.9 & -- & -- & 58.2 & -- \\
\midrule
\rowcolor{gray!15}\multicolumn{7}{c}{\textbf{Open-source MLLMs Method}} \\ 
TimeChat~\cite{ren2024timechat}  & 32.2 & 13.4 & 32.2 & 36.2 & 20.2 & 21.8  \\
VTimeLLM~\cite{huang2024vtimellm}  & 27.5 & 11.4 & 31.2 & 44.0 & 27.8 & 30.4  \\
VideoChat-Flash~\cite{li2024videochatflash} & 53.1 & 27.6 & -- & -- & -- & -- \\
TRACE~\cite{guo2024trace} & 61.7 & 41.4 & 41.4 & 37.7 & 24.0 & 39.0 \\
HawkEye~\cite{wang2024hawkeye}  & 58.3 & 28.8 & -- & 55.9 & 34.7 & --  \\
TimeSuite~\cite{zeng2025timesuite}  & 67.1 & 43.0 & -- & -- & -- & --  \\
Time-R1~\cite{wang2025timer1posttraininglargevision}  & 72.2 & 50.1 & -- & 58.6 & 39.0 & --  \\
DeepVideo-R1-7B~\cite{park2025deepvideor1videoreinforcementfinetuning} & 71.7 & 50.6 & \textbf{61.2} & 33.9 & 18.0 & 36.9 \\
VideoChat-R1-7B~\cite{li2025videochatr1enhancingspatiotemporalperception} & 71.7 & 50.2 & 60.8 & 33.4 & 17.7 & 36.6 \\
TimeZero-7B~\cite{wang2025timer1posttraininglargevision} & 60.8 & 35.3 & 58.1 & 39.0 & 21.4 & 40.5 \\
Temporal-RLT-7B\cite{li2025reinforcementlearningtuningvideollms} & 67.9 & 44.1 & 57.0 & 38.4 & 20.2 & 39.0 \\

\midrule
\rowcolor{gray!20}
\textbf{TimeScope-7B} & \textbf{78.9} & \textbf{61.2} & 56.2 & \textbf{66.9} & \textbf{56.0} & \textbf{46.0} \\
\hline
\end{tabular}
\vspace{-8px} 
\caption{Performance comparison on short video temporal grounding tasks including Charades-STA and ActivityNet. }
\label{tab:compariso}
\end{table*}

\textbf{Progressive Reasoning.}
With both $HVR$ and $FVR$ prepared, TimeScope performs grounding in a coarse-to-fine manner. In the first step, only the compact holistic representation $HVR$ is kept in GPU memory, while the high-resolution $FVR$ remains stored in CPU memory. Leveraging the lightweight long-range context encoded in $HVR$, the MLLM efficiently predicts a coarse temporal interval that is most likely to contain the task-relevant moment for the query $Q_{\text{task}}$. This stage effectively eliminates the majority of irrelevant frames and substantially reduces the temporal search space.
Next, TimeScope reloads only the fine-grained $FVR$ corresponding to the predicted interval and performs detailed reasoning. The rich local temporal and visual details preserved in $FVR$ allow the model to refine the boundaries within the coarse region and accurately localize the target moment.
By combining the global efficiency of $HVR$ with the local precision of $FVR$, the progressive reasoning process achieves high localization accuracy while maintaining low computational cost of processing the long video at high resolution.

\subsection{ToTG-Pile}
\label{sec:dataset}

To maximize TimeScope's capacity for task-oriented temporal grounding, we created \textbf{ToTG-Pile}---a large-scale dataset bridging conventional temporal grounding data with Task-oriented Temporal Grounding requirements. ToTG-Pile features two key characteristics: (1) all samples follow task-oriented query definitions (Sec.~\ref{sec:problem}), where queries are formulated as task instructions rather than explicit temporal descriptions, and (2) each sample includes both ground-truth temporal intervals and chain-of-thought (CoT) annotations that detail the localization reasoning process.

We constructed the dataset by collecting diverse videos with broad coverage of tasks and visual contexts. Specifically, we leverage comprehensive Video VQA datasets such as VideoR1~\cite{feng2025video} and FineVideo~\cite{Farré2024FineVideo}. To curate high-quality reasoning data, we employ a three-stage pipeline: (1) \textbf{Answer-aware prompting}: we feed answers from original VQA samples into expert temporal grounding models to obtain candidate temporal intervals; (2) \textbf{Cross-validation filtering}: we leverage multiple expert models for cross-validation, discarding low-quality samples with IoU $<$ 0.1 across models; (3) \textbf{CoT annotation generation}: we use reasoning-capable MLLMs to generate chain-of-thought annotations that detail the localization reasoning process. Additionally, to extend the dataset to long-video scenarios, we synthesized 90K long clips (10 minutes each) by concatenating short videos and generated corresponding temporal grounding queries from video captions. This augmentation enhances TimeScope's capability in handling both conventional temporal grounding and task-oriented temporal grounding in extended video contexts. 

Overall, ToTG-Pile provides a large-scale, diverse, and reasoning-oriented foundation for training MLLMs to perform effective and generalizable task-oriented temporal grounding in long videos. 

\section{Experiment}
\subsection{{Implementation details}}

We adopt VideoXL-2~\cite{qin2025videoxl2longvideounderstandingtaskaware} as our backbone for two reasons: (1) it can process very long video sequences, enabling straightforward construction of long-video temporal understanding methods, and (2) its internal design interleaves timestamp tokens, providing the model with a strong built-in temporal awareness. \noindent\textbf{Stage 1: Basic Localization.} We use the temporal grounding splits of ToTG-Pile and train the model to predict target time intervals directly from raw video and task descriptions, bootstrapping its basic localization ability. \noindent\textbf{Stage 2: Coarse-to-Fine Refinement.} We apply heavy temporal augmentations (random cropping, shifting, and scaling of time spans) to training videos, forcing the model to first estimate a coarse temporal window from abstract video representations and then refine it into fine-grained intervals using detailed representations.

\subsection{{Results on Task-oriented temporal grounding}}

\noindent\textbf{Evaluation on ToTG-Bench.} We evaluate TimeScope on ToTG-Bench, which categorizes videos into three duration segments: short ($<$180s), medium (180--600s), and long ($>$600s). We report IoU@0.3 and IoU@0.5 metrics and compare TimeScope against both specialized temporal understanding MLLMs and general video understanding models. As shown in Table~\ref{tab:lvtgbench}, TimeScope demonstrates strong and consistent performance across all video durations. Most notably, TimeScope outperforms all baselines by 20--30 points on medium and long videos, demonstrating substantial advantages in handling task-oriented temporal grounding in extended contexts. On short videos, TimeScope achieves competitive performance, trailing the state-of-the-art Temporal-RLT~\cite{li2025reinforcementlearningtuningvideollms} by less than 2 points while significantly surpassing it on longer videos. These results highlight TimeScope's effectiveness in task-oriented temporal localization, particularly its ability to perform accurate reasoning and grounding in long-video scenarios while maintaining robust performance across different video durations.

 We further evaluate TimeScope and baseline models on ToTG-Bench across different task categories using IoU@0.5 metric, as shown in Table~\ref{tab:main_result_1}. TimeScope demonstrates strong performance across all task categories. Notably, in reasoning tasks (including Object Reasoning, Spatial Reasoning, Temporal Reasoning, and Action Reasoning), TimeScope outperforms all baselines by 10--20 points, highlighting its superior capability in task-oriented reasoning and temporal localization. Across other task categories, TimeScope consistently achieves top-tier performance, demonstrating robust generalization to diverse task types.


\begin{table}[h]
\centering
\fontsize{8}{10}\selectfont 
\begin{tabular}{lcc}
\toprule
\textbf{Model} & \textbf{R1@0.5} & \textbf{R1@0.7} \\
\midrule
Qwen2.5-VL-7B & 0.0 & 0.0 \\
UniTime & 62.9 & 62.9 \\
Keye-1.5-VL-8B & 63.9 & 49.1 \\
\midrule
\rowcolor{gray!20}
\textbf{TimeScope-7B} & \textbf{87.5} & \textbf{85.2} \\
\bottomrule
\end{tabular}
\vspace{-8px} 
\caption{Performance comparison on long video temporal grounding benchmark V-StaR (duration$>$300).}
\label{tab:vstar}
\end{table}

\subsection{Results on Traditional Temporal Grounding}
We conduct a comprehensive comparison of TimeScope against traditional and MLLM-based methods on conventional temporal grounding benchmarks, covering both short-video and long-video settings.

\noindent\textbf{Short-video Benchmarks.} As shown in Table~\ref{tab:compariso}, TimeScope achieves state-of-the-art performance across all short-video benchmarks. On Charades-STA, TimeScope attains an R1@0.7 score of 64.0, significantly surpassing VideoChat-Flash (27.6), TimeSuite (43.0), and Time-R1 (50.1). On ActivityNet, it achieves an R1@0.7 score of 59.0, outperforming HawkEye (34.7) and Time-R1 (39.0). Notably, TimeScope maintains a smaller gap between R1@0.5 and R1@0.7 compared to most baselines, indicating its capability for more precise temporal localization.

\noindent\textbf{Long-video Benchmarks.} As shown in Table~\ref{tab:vstar}, TimeScope achieves an R1@0.7 score of 85.2 on V-STaR (with videos up to 300 seconds), substantially exceeding UniTime (62.9) and Keye-1.5-VL (49.1). This demonstrates TimeScope's strong capability in handling extended temporal grounding tasks.

\begin{table}[h]
\centering
\fontsize{8}{10}\selectfont 
\resizebox{0.5\textwidth}{!}{
\begin{tabular}{l|c|c|c}
\hline
\multirow{2}{*}{\textbf{Method}}  & \textbf{CG-Bench} & \textbf{MLVU} & \textbf{LongVideoBench} \\
  & \textbf{Acc.} & \textbf{Acc.} & \textbf{Acc.} \\
\midrule
Uniform Sample  & 33.87 & 60.53 & 54.82 \\
UniVTG~\cite{lin2023univtg}   & 34.87 & 62.56 & 54.67 \\
VTimeLLM~\cite{huang2024vtimellm} & 34.60 & 59.52 & 54.30 \\
TimeSuite~\cite{zeng2025timesuite}   & 32.47 & 58.51 & 53.25 \\
UniTime-Full~\cite{li2025universalvideotemporalgrounding} & \textbf{40.30} & 66.50 & 56.47 \\
\midrule
\rowcolor{gray!20}
\textbf{TimeScope-7B}  & 38.47 & \textbf{68.12} & \textbf{58.34} \\
\hline
\end{tabular}
}
\vspace{-8px} 
\caption{Performance comparison on Long Video Understanding tasks including CG-Bench, MLVU and LongVideoBench. }
\label{tab:videoqa}
\end{table}

\subsection{Benefits to Long-Video Understanding}
As discussed in Section~\ref{sec:problem}, TimeScope's strong performance on task-oriented temporal grounding demonstrates its potential to help MLLMs capture critical information in long videos for question answering. To validate this, we conduct experiments where TimeScope and baseline temporal grounding models first localize relevant time intervals, and then feed frames from the predicted intervals into Qwen2-VL-7B~\cite{bai2025qwen25vltechnicalreport} for answer generation. We compare these results against other grounding models and a default uniform sampling baseline without temporal grounding.

We evaluate on three long-video understanding benchmarks: CG-Bench, MLVU, and LongVideoBench. As shown in Table~\ref{tab:videoqa}, TimeScope demonstrates strong performance and brings substantial improvements over uniform sampling across all benchmarks, surpassing most temporal grounding baselines. On MLVU and LongVideoBench, TimeScope achieves the highest scores of 68.12 and 58.34 respectively, demonstrating its effectiveness in identifying task-relevant temporal segments for long-video understanding. Notably, while many video understanding questions require information beyond a single temporal segment, TimeScope's task-oriented grounding still provides meaningful performance gains, validating its practical utility for downstream applications.

\subsection{Ablation Studies}
\textbf{Effectiveness of Progressive Reasoning.}
To evaluate the effectiveness and necessity of progressive reasoning in long-video scenarios, we conducted an ablation study comparing two settings: progressive reasoning versus standard single prediction. The experiment was performed on samples longer than 300 seconds from the V-STaR benchmark and samples exceeding 600 seconds from the ToTG-Bench, with results presented in Table~\ref{tab:twostage}. The findings demonstrate that progressive reasoning achieves substantial improvements over the single-step baseline across both traditional temporal grounding tasks and task-oriented tasks in long videos. Particularly notable is the performance gain huge improvement in IoU@0.7, which confirms the effectiveness of progressively narrowing the search space for achieving precise temporal localization.

\textbf{Efficiency of Progressive Reasoning.}
Despite employing progressive reasoning to achieve higher prediction accuracy, TimeScope remains highly efficient—thanks to its streaming-based video representation construction and carefully designed dual-level representation. As shown in Table~\ref{tab:efficiency}, our framework enables the MLLM backbone to process significantly more frames while simultaneously improving throughput in both the prefill and decode stages, outperforming baseline methods without TimeScope in both speed and scalability.

\begin{table}[t]
\centering
\fontsize{8}{10}\selectfont 
\begin{tabular}{lcc|cc}
\toprule
\multirow{2}{*}{\textbf{Method}} & \multicolumn{2}{c|}{\textbf{V-STaR(long)}} & \multicolumn{2}{c}{\textbf{ToTG-bench(long)}} \\
 & R1@0.5 & R1@0.7 & R1@0.5 & R1@0.7\\
\midrule

Zeroshot  & 73.2 & 61.4 & 37.8 & 29.5\\
\midrule
\rowcolor{gray!20}

Progressive Reasoning & \textbf{86.4} & \textbf{83.0} & \textbf{37.8} & \textbf{34.0} \\

\bottomrule
\end{tabular}%
\vspace{-8px} 
\caption{Comparison of progressive reasoning versus standard prediction on long-video subsets of V-STaR (duration $>$ 300s) and ToTG-Bench.}
\label{tab:twostage}
\end{table}


\textbf{Effectiveness of Holistic Video Representation.}
As outlined in Sec.~\ref{sec:framework}, TimeScope enables dense frame sampling for long videos—made feasible by the efficient design of its Holistic Video Representation (HVR). As demonstrated in Table~\ref{tab:holistic}, with HVR, TimeScope supports dense sampling even for extremely long videos (up to 2,000 frames). This capability directly translates into significant gains in temporal grounding precision, particularly for fine-grained event localization.

\begin{table}[t]
\centering
\fontsize{8}{10}\selectfont 
\resizebox{0.4\textwidth}{!}{
\begin{tabular}{c|c|cc|c}
\toprule
\textbf{Method} & \textbf{Frames} &  \textbf{Prefill} &  \textbf{Decode} & \textbf{Sum} \\
\midrule
w/o TimeScope & 800 & 2903ms & 816ms & 3719ms\\
\midrule

\multirow{2}{*}{w TimeScope} & 800  & 2259ms & 517ms & 2776ms\\
                             & 1200 & 3205ms & 627ms & 3832ms\\

\bottomrule
\end{tabular}
}
\vspace{-8px} 
\caption{The efficiency of TimeScope.}
\label{tab:efficiency}
\end{table}

\textbf{Efficiency of ToTG-pile Dataset.}
To further analyze the impact of the ToTG-Pile dataset, we retrained the model by excluding it from the training set and reported the corresponding results on short-video temporal benchmark Charades-STA, long-video benchmarks V-STaR and ToTG-bench (long), as shown in Table~\ref{tab:totgpile}. It can be observed that the performance gap becomes particularly pronounced on V-STaR and ToTG-bench (long), where ToTG-pile brings performance gains of 40 and 20 points, respectively, compared to the setting without ToTG-pile. This demonstrates the crucial value of ToTG-pile in establishing TimeScope's long-video understanding capability and task-oriented reasoning proficiency.

\begin{table}[!t]
\centering
\fontsize{8}{10}\selectfont

\begin{subtable}{0.48\textwidth}
\centering
\resizebox{\textwidth}{!}{
\begin{tabular}{lccc|cc}
\toprule
\multirow{2}{*}{\textbf{Setting}} & 
\multirow{2}{*}{\textbf{Max Frames}} & 
\multicolumn{2}{c|}{\textbf{V-STaR(long)}} & \multicolumn{2}{c}{\textbf{ToTG-Bench(long)}} \\
&  & R1@0.5 & R1@0.7 & R1@0.5 & R1@0.7\\
\midrule
w/o Holistic VR. & 800 & 73.2 & 61.4 & 37.8 & 29.5 \\
\midrule
\rowcolor{gray!20}
w Holistic VR. & 2000 & \textbf{80.7} & \textbf{76.1} & \textbf{41.0} & \textbf{37.1} \\
\bottomrule
\end{tabular}
}
\caption{The effectiveness of Holistic Video Representation.}
\vspace{5px} 
\label{tab:holistic}
\end{subtable}
\hfill

\begin{subtable}{0.48\textwidth}
\centering
\setlength{\tabcolsep}{1.5pt}
\begin{tabular}{lcc|cc|cc}
\toprule
\multirow{2}{*}{\textbf{Method}} & \multicolumn{2}{c|}{\textbf{Charades-STA}}& \multicolumn{2}{c|}{\textbf{V-STaR(long)}} & \multicolumn{2}{c}{\textbf{ToTG-bench(long)}} \\
 & R1@0.5 & R1@0.7 & R1@0.5 & R1@0.7 & R1@0.5 & R1@0.7\\
\midrule
W/o ToTG-Pile & 74.9 & 55.8 & 47.7 & 39.7 & 29.4 & 22.7\\
\midrule
\rowcolor{gray!20}
W ToTG-Pile & \textbf{78.9} & \textbf{61.2} & \textbf{86.4} & \textbf{83.0} & \textbf{37.8} & \textbf{34.0} \\
\bottomrule
\end{tabular}
\caption{Analysis of training effect from ToTG-Pile.}
\label{tab:totgpile}
\end{subtable}

\caption{Ablation studies on model components.}
\label{tab:combined}
\end{table}
\section{Qualitative results}
We show the results of TimeScope in Figure \ref{fig:definition}, which can be seen that TimeScope exhibits robust performance with good coarse-grained segment retrieval and
fine-grained temporal grounding capabilities. More results can be seen in Appendix.

\begin{figure}[]
    \centering
    \includegraphics[width=1.0\linewidth]{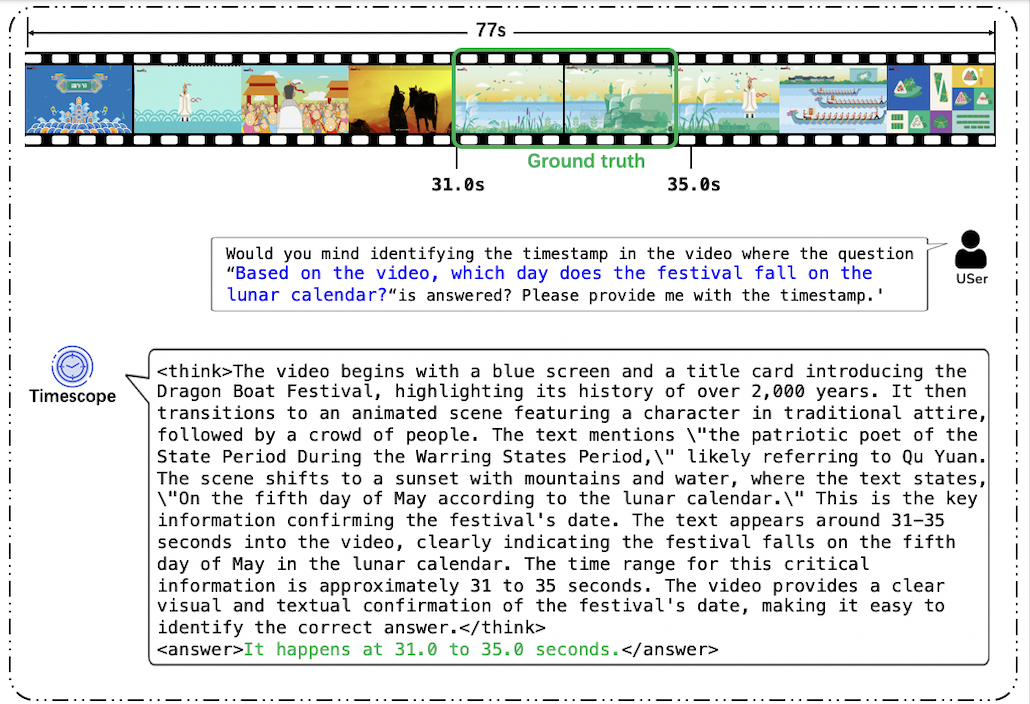}
    \caption{Visualization of TimeScope.}
    \label{fig:definition}
\end{figure} 

\section{Conclusion}
In this work, we define a new task—Task-Oriented Temporal Grounding (ToTG)—and formally conceptualize the aforementioned challenges.
To foster research in this emerging area, we introduce ToTG-Bench, a benchmark designed to evaluate temporal grounding performance on diverse, real-world, long-form video-understanding scenarios. To tackle these challenges, we propose TimeScope, a novel framework that solves ToTG through step-by-step reasoning. To strengthen TimeScope, we release ToTG-Pile, a dataset expressly engineered to optimize MLLMs for task-oriented temporal grounding. Harvested from diverse real-world long-video corpora and annotated via a carefully engineered pipeline, ToTG-Pile provides large-scale, high-quality training data. Extensive experiments across a wide spectrum of settings show that TimeScope achieves substantial improvements over existing methods on both traditional benchmarks and ToTG-Bench.
We hope this work will stimulate future research on Task-Oriented Temporal Grounding and propel MLLMs toward deeper temporal understanding of video.
\vspace{0.8cm}

{
    \small
    \bibliographystyle{ieeenat_fullname}
    \bibliography{main}

\begin{thebibliography}{49}
\providecommand{\natexlab}[1]{#1}
\providecommand{\url}[1]{\texttt{#1}}
\expandafter\ifx\csname urlstyle\endcsname\relax
  \providecommand{\doi}[1]{doi: #1}\else
  \providecommand{\doi}{doi: \begingroup \urlstyle{rm}\Url}\fi

\bibitem[Bai et~al.(2025)Bai, Chen, Liu, Wang, Ge, Song, Dang, Wang, Wang, Tang, Zhong, Zhu, Yang, Li, Wan, Wang, Ding, Fu, Xu, Ye, Zhang, Xie, Cheng, Zhang, Yang, Xu, and Lin]{bai2025qwen25vltechnicalreport}
Shuai Bai, Keqin Chen, Xuejing Liu, Jialin Wang, Wenbin Ge, Sibo Song, Kai Dang, Peng Wang, Shijie Wang, Jun Tang, Humen Zhong, Yuanzhi Zhu, Mingkun Yang, Zhaohai Li, Jianqiang Wan, Pengfei Wang, Wei Ding, Zheren Fu, Yiheng Xu, Jiabo Ye, Xi Zhang, Tianbao Xie, Zesen Cheng, Hang Zhang, Zhibo Yang, Haiyang Xu, and Junyang Lin.
\newblock Qwen2.5-vl technical report, 2025.

\bibitem[Caba~Heilbron et~al.(2015)Caba~Heilbron, Escorcia, Ghanem, and Carlos~Niebles]{caba2015activitynet}
Fabian Caba~Heilbron, Victor Escorcia, Bernard Ghanem, and Juan Carlos~Niebles.
\newblock Activitynet: A large-scale video benchmark for human activity understanding.
\newblock In \emph{Proceedings of the ieee conference on computer vision and pattern recognition}, pages 961--970, 2015.

\bibitem[Chen et~al.(2024)Chen, Liu, Huang, He, Pei, Xu, Wang, Lu, and Wang]{chen2024cg}
Guo Chen, Yicheng Liu, Yifei Huang, Yuping He, Baoqi Pei, Jilan Xu, Yali Wang, Tong Lu, and Limin Wang.
\newblock Cg-bench: Clue-grounded question answering benchmark for long video understanding.
\newblock \emph{arXiv preprint arXiv:2412.12075}, 2024.

\bibitem[Chen et~al.(2025)Chen, Li, Wang, Jiang, Liu, Lu, Huang, Byeon, Le, Rintamaki, Poon, Ehrlich, Rintamaki, Poon, Lu, Wang, Catanzaro, Kautz, Tao, Yu, and Liu]{chen2025eagle25boostinglongcontext}
Guo Chen, Zhiqi Li, Shihao Wang, Jindong Jiang, Yicheng Liu, Lidong Lu, De-An Huang, Wonmin Byeon, Matthieu Le, Tuomas Rintamaki, Tyler Poon, Max Ehrlich, Tuomas Rintamaki, Tyler Poon, Tong Lu, Limin Wang, Bryan Catanzaro, Jan Kautz, Andrew Tao, Zhiding Yu, and Guilin Liu.
\newblock Eagle 2.5: Boosting long-context post-training for frontier vision-language models, 2025.

\bibitem[Cheng et~al.(2025{\natexlab{a}})Cheng, Hu, Liu, Si, Li, and Gong]{cheng2025v}
Zixu Cheng, Jian Hu, Ziquan Liu, Chenyang Si, Wei Li, and Shaogang Gong.
\newblock V-star: Benchmarking video-llms on video spatio-temporal reasoning.
\newblock \emph{arXiv preprint arXiv:2503.11495}, 2025{\natexlab{a}}.

\bibitem[Cheng et~al.(2025{\natexlab{b}})Cheng, Hu, Liu, Si, Li, and Gong]{cheng2025vstarbenchmarkingvideollmsvideo}
Zixu Cheng, Jian Hu, Ziquan Liu, Chenyang Si, Wei Li, and Shaogang Gong.
\newblock V-star: Benchmarking video-llms on video spatio-temporal reasoning, 2025{\natexlab{b}}.

\bibitem[Devlin et~al.(2019)Devlin, Chang, Lee, and Toutanova]{devlin2019bertpretrainingdeepbidirectional}
Jacob Devlin, Ming-Wei Chang, Kenton Lee, and Kristina Toutanova.
\newblock Bert: Pre-training of deep bidirectional transformers for language understanding, 2019.

\bibitem[Farré et~al.(2024)Farré, Marafioti, Tunstall, Von~Werra, and Wolf]{Farré2024FineVideo}
Miquel Farré, Andi Marafioti, Lewis Tunstall, Leandro Von~Werra, and Thomas Wolf.
\newblock Finevideo, 2024.

\bibitem[Feng et~al.(2025)Feng, Gong, Li, Guo, Wang, Peng, Wu, Zhang, Wang, and Yue]{feng2025video}
Kaituo Feng, Kaixiong Gong, Bohao Li, Zonghao Guo, Yibing Wang, Tianshuo Peng, Junfei Wu, Xiaoying Zhang, Benyou Wang, and Xiangyu Yue.
\newblock Video-r1: Reinforcing video reasoning in mllms.
\newblock \emph{arXiv preprint arXiv:2503.21776}, 2025.

\bibitem[Fu et~al.(2025)Fu, Dai, Luo, Li, Ren, Zhang, Wang, Zhou, Shen, Zhang, et~al.]{fu2025video}
Chaoyou Fu, Yuhan Dai, Yongdong Luo, Lei Li, Shuhuai Ren, Renrui Zhang, Zihan Wang, Chenyu Zhou, Yunhang Shen, Mengdan Zhang, et~al.
\newblock Video-mme: The first-ever comprehensive evaluation benchmark of multi-modal llms in video analysis.
\newblock In \emph{Proceedings of the Computer Vision and Pattern Recognition Conference}, pages 24108--24118, 2025.

\bibitem[Gordeev et~al.(2024)Gordeev, Dokholyan, Tolstykh, and Kuprashevich]{gordeev2024saliency}
Aleksandr Gordeev, Vladimir Dokholyan, Irina Tolstykh, and Maksim Kuprashevich.
\newblock Saliency-guided detr for moment retrieval and highlight detection.
\newblock \emph{arXiv preprint arXiv:2410.01615}, 2024.

\bibitem[Guo et~al.(2024)Guo, Liu, Li, Liu, Chen, and Tang]{guo2024trace}
Yongxin Guo, Jingyu Liu, Mingda Li, Qingbin Liu, Xi Chen, and Xiaoying Tang.
\newblock Trace: Temporal grounding video llm via causal event modeling.
\newblock \emph{arXiv preprint arXiv:2410.05643}, 2024.

\bibitem[Huang et~al.(2024{\natexlab{a}})Huang, Wang, Chen, Song, and Zhu]{huang2024vtimellm}
Bin Huang, Xin Wang, Hong Chen, Zihan Song, and Wenwu Zhu.
\newblock Vtimellm: Empower llm to grasp video moments.
\newblock In \emph{Proceedings of the IEEE/CVF Conference on Computer Vision and Pattern Recognition}, pages 14271--14280, 2024{\natexlab{a}}.

\bibitem[Huang et~al.(2024{\natexlab{b}})Huang, Liao, Radhakrishnan, Yin, Molchanov, Yu, and Kautz]{huang2024lita}
De-An Huang, Shijia Liao, Subhashree Radhakrishnan, Hongxu Yin, Pavlo Molchanov, Zhiding Yu, and Jan Kautz.
\newblock Lita: Language instructed temporal-localization assistant.
\newblock In \emph{European Conference on Computer Vision}, pages 202--218. Springer, 2024{\natexlab{b}}.

\bibitem[Huang et~al.(2025)Huang, Radhakrishnan, Yu, and Kautz]{huang2025frag}
De-An Huang, Subhashree Radhakrishnan, Zhiding Yu, and Jan Kautz.
\newblock Frag: Frame selection augmented generation for long video and long document understanding.
\newblock \emph{arXiv preprint arXiv:2504.17447}, 2025.

\bibitem[Jang et~al.(2023)Jang, Park, Kim, Kwon, and Sohn]{jang2023knowing}
Jinhyun Jang, Jungin Park, Jin Kim, Hyeongjun Kwon, and Kwanghoon Sohn.
\newblock Knowing where to focus: Event-aware transformer for video grounding.
\newblock In \emph{Proceedings of the IEEE/CVF International Conference on Computer Vision}, pages 13846--13856, 2023.

\bibitem[Lei et~al.(2021)Lei, Berg, and Bansal]{lei2021detecting}
Jie Lei, Tamara~L Berg, and Mohit Bansal.
\newblock Detecting moments and highlights in videos via natural language queries.
\newblock \emph{Advances in Neural Information Processing Systems}, 34:\penalty0 11846--11858, 2021.

\bibitem[Li et~al.(2025{\natexlab{a}})Li, Han, Liao, Luo, Gao, Yan, and Liu]{li2025reinforcementlearningtuningvideollms}
Hongyu Li, Songhao Han, Yue Liao, Junfeng Luo, Jialin Gao, Shuicheng Yan, and Si Liu.
\newblock Reinforcement learning tuning for videollms: Reward design and data efficiency, 2025{\natexlab{a}}.

\bibitem[Li et~al.(2024)Li, Wang, Yu, Zeng, Zhu, Huang, Gao, Li, He, Wang, Qiao, Wang, and Wang]{li2024videochatflash}
Xinhao Li, Yi Wang, Jiashuo Yu, Xiangyu Zeng, Yuhan Zhu, Haian Huang, Jianfei Gao, Kunchang Li, Yinan He, Chenting Wang, Yu Qiao, Yali Wang, and Limin Wang.
\newblock Videochat-flash: Hierarchical compression for long-context video modeling.
\newblock \emph{arXiv preprint arXiv:2501.00574}, 2024.

\bibitem[Li et~al.(2025{\natexlab{b}})Li, Wang, Yu, Zeng, Zhu, Huang, Gao, Li, He, Wang, Qiao, Wang, and Wang]{li2025videochatflashhierarchicalcompressionlongcontext}
Xinhao Li, Yi Wang, Jiashuo Yu, Xiangyu Zeng, Yuhan Zhu, Haian Huang, Jianfei Gao, Kunchang Li, Yinan He, Chenting Wang, Yu Qiao, Yali Wang, and Limin Wang.
\newblock Videochat-flash: Hierarchical compression for long-context video modeling, 2025{\natexlab{b}}.

\bibitem[Li et~al.(2025{\natexlab{c}})Li, Yan, Meng, Dong, Zeng, He, Wang, Qiao, Wang, and Wang]{li2025videochatr1enhancingspatiotemporalperception}
Xinhao Li, Ziang Yan, Desen Meng, Lu Dong, Xiangyu Zeng, Yinan He, Yali Wang, Yu Qiao, Yi Wang, and Limin Wang.
\newblock Videochat-r1: Enhancing spatio-temporal perception via reinforcement fine-tuning, 2025{\natexlab{c}}.

\bibitem[Li et~al.(2025{\natexlab{d}})Li, Di, Zhai, Huang, Wang, and Xie]{li2025universal}
Zeqian Li, Shangzhe Di, Zhonghua Zhai, Weilin Huang, Yanfeng Wang, and Weidi Xie.
\newblock Universal video temporal grounding with generative multi-modal large language models.
\newblock \emph{arXiv preprint arXiv:2506.18883}, 2025{\natexlab{d}}.

\bibitem[Li et~al.(2025{\natexlab{e}})Li, Di, Zhai, Huang, Wang, and Xie]{li2025universalvideotemporalgrounding}
Zeqian Li, Shangzhe Di, Zhonghua Zhai, Weilin Huang, Yanfeng Wang, and Weidi Xie.
\newblock Universal video temporal grounding with generative multi-modal large language models, 2025{\natexlab{e}}.

\bibitem[Lin et~al.(2023)Lin, Zhang, Chen, Pramanick, Gao, Wang, Yan, and Shou]{lin2023univtg}
Kevin~Qinghong Lin, Pengchuan Zhang, Joya Chen, Shraman Pramanick, Difei Gao, Alex~Jinpeng Wang, Rui Yan, and Mike~Zheng Shou.
\newblock Univtg: Towards unified video-language temporal grounding.
\newblock In \emph{Proceedings of the IEEE/CVF International Conference on Computer Vision}, pages 2794--2804, 2023.

\bibitem[Moon et~al.(2023{\natexlab{a}})Moon, Hyun, Lee, and Heo]{moon2023correlation}
WonJun Moon, Sangeek Hyun, SuBeen Lee, and Jae-Pil Heo.
\newblock Correlation-guided query-dependency calibration for video temporal grounding.
\newblock \emph{arXiv preprint arXiv:2311.08835}, 2023{\natexlab{a}}.

\bibitem[Moon et~al.(2023{\natexlab{b}})Moon, Hyun, Park, Park, and Heo]{moon2023query}
WonJun Moon, Sangeek Hyun, SangUk Park, Dongchan Park, and Jae-Pil Heo.
\newblock Query-dependent video representation for moment retrieval and highlight detection.
\newblock In \emph{Proceedings of the IEEE/CVF conference on computer vision and pattern recognition}, pages 23023--23033, 2023{\natexlab{b}}.

\bibitem[Mu et~al.(2024)Mu, Mo, and Li]{CVPR2024SnAG}
Fangzhou Mu, Sicheng Mo, and Yin Li.
\newblock Snag: Scalable and accurate video grounding.
\newblock \emph{2024 IEEE/CVF Conference on Computer Vision and Pattern Recognition (CVPR)}, pages 18930--18940, 2024.

\bibitem[Park et~al.(2025)Park, Na, Kim, and Kim]{park2025deepvideor1videoreinforcementfinetuning}
Jinyoung Park, Jeehye Na, Jinyoung Kim, and Hyunwoo~J. Kim.
\newblock Deepvideo-r1: Video reinforcement fine-tuning via difficulty-aware regressive grpo, 2025.

\bibitem[Qin et~al.(2025{\natexlab{a}})Qin, Liu, Liang, Shu, Yuan, Zhou, Xiao, Zhao, and Liu]{qin2025videoxl2longvideounderstandingtaskaware}
Minghao Qin, Xiangrui Liu, Zhengyang Liang, Yan Shu, Huaying Yuan, Juenjie Zhou, Shitao Xiao, Bo Zhao, and Zheng Liu.
\newblock Video-xl-2: Towards very long-video understanding through task-aware kv sparsification, 2025{\natexlab{a}}.

\bibitem[Qin et~al.(2025{\natexlab{b}})Qin, Shu, Zhang, Lun, Yuan, Zhou, Xiao, Zhao, and Liu]{qin2025task}
Minghao Qin, Yan Shu, Peitian Zhang, Kun Lun, Huaying Yuan, Juenjie Zhou, Shitao Xiao, Bo Zhao, and Zheng Liu.
\newblock Task-aware kv compression for cost-effective long video understanding.
\newblock \emph{arXiv preprint arXiv:2506.21184}, 2025{\natexlab{b}}.

\bibitem[Radford et~al.(2021)Radford, Kim, Hallacy, Ramesh, Goh, Agarwal, Sastry, Askell, Mishkin, Clark, Krueger, and Sutskever]{radford2021learningtransferablevisualmodels}
Alec Radford, Jong~Wook Kim, Chris Hallacy, Aditya Ramesh, Gabriel Goh, Sandhini Agarwal, Girish Sastry, Amanda Askell, Pamela Mishkin, Jack Clark, Gretchen Krueger, and Ilya Sutskever.
\newblock Learning transferable visual models from natural language supervision, 2021.

\bibitem[Ren et~al.(2024)Ren, Yao, Li, Sun, and Hou]{ren2024timechat}
Shuhuai Ren, Linli Yao, Shicheng Li, Xu Sun, and Lu Hou.
\newblock Timechat: A time-sensitive multimodal large language model for long video understanding.
\newblock In \emph{Proceedings of the IEEE/CVF Conference on Computer Vision and Pattern Recognition}, pages 14313--14323, 2024.

\bibitem[Sigurdsson et~al.(2016)Sigurdsson, Varol, Wang, Farhadi, Laptev, and Gupta]{sigurdsson2016Charades}
Gunnar~A Sigurdsson, G{\"u}l Varol, Xiaolong Wang, Ali Farhadi, Ivan Laptev, and Abhinav Gupta.
\newblock Hollywood in homes: Crowdsourcing data collection for activity understanding.
\newblock In \emph{Proceedings of the European Conference on Computer Vision (ECCV)}, 2016.

\bibitem[Song et~al.(2024{\natexlab{a}})Song, Chai, Wang, Zhang, Zhou, Wu, Chi, Guo, Ye, Zhang, Lu, Hwang, and Wang]{song2024moviechatdensetokensparse}
Enxin Song, Wenhao Chai, Guanhong Wang, Yucheng Zhang, Haoyang Zhou, Feiyang Wu, Haozhe Chi, Xun Guo, Tian Ye, Yanting Zhang, Yan Lu, Jenq-Neng Hwang, and Gaoang Wang.
\newblock Moviechat: From dense token to sparse memory for long video understanding, 2024{\natexlab{a}}.

\bibitem[Song et~al.(2024{\natexlab{b}})Song, Chai, Ye, Hwang, Li, and Wang]{song2024moviechatquestionawaresparsememory}
Enxin Song, Wenhao Chai, Tian Ye, Jenq-Neng Hwang, Xi Li, and Gaoang Wang.
\newblock Moviechat+: Question-aware sparse memory for long video question answering, 2024{\natexlab{b}}.

\bibitem[Wang et~al.(2025{\natexlab{a}})Wang, Chen, Huang, Li, Li, Li, Alvarez, Zhang, and Yu]{wang2025videoitg}
Shihao Wang, Guo Chen, De-an Huang, Zhiqi Li, Minghan Li, Guilin Li, Jose~M Alvarez, Lei Zhang, and Zhiding Yu.
\newblock Videoitg: Multimodal video understanding with instructed temporal grounding.
\newblock \emph{arXiv preprint arXiv:2507.13353}, 2025{\natexlab{a}}.

\bibitem[Wang et~al.(2024)Wang, Meng, Liang, Wang, Liu, and Zhao]{wang2024hawkeye}
Yueqian Wang, Xiaojun Meng, Jianxin Liang, Yuxuan Wang, Qun Liu, and Dongyan Zhao.
\newblock Hawkeye: Training video-text llms for grounding text in videos, 2024.

\bibitem[Wang et~al.(2025{\natexlab{b}})Wang, Wang, Xu, Du, Lin, Xiao, Yue, Ju, Zhang, Yang, Fang, He, Luo, Wang, Lin, Luan, and Jin]{wang2025timer1posttraininglargevision}
Ye Wang, Ziheng Wang, Boshen Xu, Yang Du, Kejun Lin, Zihan Xiao, Zihao Yue, Jianzhong Ju, Liang Zhang, Dingyi Yang, Xiangnan Fang, Zewen He, Zhenbo Luo, Wenxuan Wang, Junqi Lin, Jian Luan, and Qin Jin.
\newblock Time-r1: Post-training large vision language model for temporal video grounding, 2025{\natexlab{b}}.

\bibitem[Wu et~al.(2024)Wu, Li, Chen, and Li]{wu2024longvideobench}
Haoning Wu, Dongxu Li, Bei Chen, and Junnan Li.
\newblock Longvideobench: A benchmark for long-context interleaved video-language understanding.
\newblock \emph{Advances in Neural Information Processing Systems}, 37:\penalty0 28828--28857, 2024.

\bibitem[Xiao et~al.(2024)Xiao, Yao, Li, and Chua]{xiao2024can}
Junbin Xiao, Angela Yao, Yicong Li, and Tat-Seng Chua.
\newblock Can i trust your answer? visually grounded video question answering.
\newblock In \emph{Proceedings of the IEEE/CVF Conference on Computer Vision and Pattern Recognition}, pages 13204--13214, 2024.

\bibitem[Yang et~al.(2025{\natexlab{a}})Yang, Wen, Ding, Liu, Chu, Song, Rao, Yi, Li, Zang, Yang, Zhou, Zhang, Shen, Peng, Ding, Wang, Fan, Ju, Huang, Cao, Chen, Hua, Chen, Jiang, Tang, Gai, Wei, Wang, Wang, Na, Zhang, Mao, Huang, Zhang, Gao, Chen, Yuan, Wu, Hu, Lu, Zhang, Yang, Chen, Lu, Wu, Ling, Yang, Li, Xu, Gao, Li, Wang, Ren, Hu, Wang, Wang, Luo, Li, Hu, and Zhang]{yang2025kwaikeyevl15technical}
Biao Yang, Bin Wen, Boyang Ding, Changyi Liu, Chenglong Chu, Chengru Song, Chongling Rao, Chuan Yi, Da Li, Dunju Zang, Fan Yang, Guorui Zhou, Guowang Zhang, Han Shen, Hao Peng, Haojie Ding, Hao Wang, Haonan Fan, Hengrui Ju, Jiaming Huang, Jiangxia Cao, Jiankang Chen, Jingyun Hua, Kaibing Chen, Kaiyu Jiang, Kaiyu Tang, Kun Gai, Muhao Wei, Qiang Wang, Ruitao Wang, Sen Na, Shengnan Zhang, Siyang Mao, Sui Huang, Tianke Zhang, Tingting Gao, Wei Chen, Wei Yuan, Xiangyu Wu, Xiao Hu, Xingyu Lu, Yi-Fan Zhang, Yiping Yang, Yulong Chen, Zeyi Lu, Zhenhua Wu, Zhixin Ling, Zhuoran Yang, Ziming Li, Di Xu, Haixuan Gao, Hang Li, Jing Wang, Lejian Ren, Qigen Hu, Qianqian Wang, Shiyao Wang, Xinchen Luo, Yan Li, Yuhang Hu, and Zixing Zhang.
\newblock Kwai keye-vl 1.5 technical report, 2025{\natexlab{a}}.

\bibitem[Yang et~al.(2025{\natexlab{b}})Yang, Wen, Ding, Liu, Chu, Song, Rao, Yi, Li, Zang, et~al.]{yang2025kwai}
Biao Yang, Bin Wen, Boyang Ding, Changyi Liu, Chenglong Chu, Chengru Song, Chongling Rao, Chuan Yi, Da Li, Dunju Zang, et~al.
\newblock Kwai keye-vl 1.5 technical report.
\newblock \emph{arXiv preprint arXiv:2509.01563}, 2025{\natexlab{b}}.

\bibitem[Yu et~al.(2024)Yu, Jin, Wang, Chen, Jin, Zuo, Xu, Sun, Zhang, Wu, et~al.]{yu2024frame}
Sicheng Yu, Chengkai Jin, Huanyu Wang, Zhenghao Chen, Sheng Jin, Zhongrong Zuo, Xiaolei Xu, Zhenbang Sun, Bingni Zhang, Jiawei Wu, et~al.
\newblock Frame-voyager: Learning to query frames for video large language models.
\newblock \emph{arXiv preprint arXiv:2410.03226}, 2024.

\bibitem[Zeng et~al.(2025)Zeng, Li, Wang, Li, Jiang, Yan, Li, Shi, Yue, Wang, Wang, Qiao, and Wang]{zeng2025timesuite}
Xiangyu Zeng, Kunchang Li, Chenting Wang, Xinhao Li, Tianxiang Jiang, Ziang Yan, Songze Li, Yansong Shi, Zhengrong Yue, Yi Wang, Yali Wang, Yu Qiao, and Limin Wang.
\newblock Timesuite: Improving {MLLM}s for long video understanding via grounded tuning.
\newblock In \emph{The Thirteenth International Conference on Learning Representations}, 2025.

\bibitem[Zhai et~al.(2023)Zhai, Mustafa, Kolesnikov, and Beyer]{zhai2023sigmoidlosslanguageimage}
Xiaohua Zhai, Basil Mustafa, Alexander Kolesnikov, and Lucas Beyer.
\newblock Sigmoid loss for language image pre-training, 2023.

\bibitem[Zhang et~al.(2020)Zhang, Peng, Fu, and Luo]{zhang20192DTAN}
Songyang Zhang, Houwen Peng, Jianlong Fu, and Jiebo Luo.
\newblock Learning 2d temporal adjacent networks for moment localization with natural language.
\newblock In \emph{Proceedings of the AAAI Conference on Artificial Intelligence}, 2020.

\bibitem[Zhou et~al.(2025)Zhou, Shu, Zhao, Wu, Liang, Xiao, Qin, Yang, Xiong, Zhang, et~al.]{zhou2025mlvu}
Junjie Zhou, Yan Shu, Bo Zhao, Boya Wu, Zhengyang Liang, Shitao Xiao, Minghao Qin, Xi Yang, Yongping Xiong, Bo Zhang, et~al.
\newblock Mlvu: Benchmarking multi-task long video understanding.
\newblock In \emph{Proceedings of the Computer Vision and Pattern Recognition Conference}, pages 13691--13701, 2025.

\bibitem[Zhu et~al.(2022)Zhu, Liu, Zhou, Di, Cheng, Yang, Xu, Xu, Wan, Sun, and Xiong]{ssrn}
Jiahao Zhu, Daizong Liu, Pan Zhou, Xing Di, Yu Cheng, Song Yang, Wenzheng Xu, Zichuan Xu, Yao Wan, Lichao Sun, and Zeyu Xiong.
\newblock Rethinking the video sampling and reasoning strategies for temporal sentence grounding.
\newblock In \emph{Findings of the Association for Computational Linguistics: EMNLP 2022}, 2022.

\bibitem[Zhu et~al.(2025)Zhu, Wang, Chen, Liu, Ye, Gu, Tian, Duan, Su, Shao, Gao, Cui, Wang, Cao, Liu, Wei, Zhang, Wang, Xu, Li, Wang, Deng, Li, He, Jiang, Luo, Wang, He, Shi, Zhang, Shao, He, Xiong, Qu, Sun, Jiao, Lv, Wu, Zhang, Deng, Ge, Chen, Wang, Dou, Lu, Zhu, Lu, Lin, Qiao, Dai, and Wang]{zhu2025internvl3exploringadvancedtraining}
Jinguo Zhu, Weiyun Wang, Zhe Chen, Zhaoyang Liu, Shenglong Ye, Lixin Gu, Hao Tian, Yuchen Duan, Weijie Su, Jie Shao, Zhangwei Gao, Erfei Cui, Xuehui Wang, Yue Cao, Yangzhou Liu, Xingguang Wei, Hongjie Zhang, Haomin Wang, Weiye Xu, Hao Li, Jiahao Wang, Nianchen Deng, Songze Li, Yinan He, Tan Jiang, Jiapeng Luo, Yi Wang, Conghui He, Botian Shi, Xingcheng Zhang, Wenqi Shao, Junjun He, Yingtong Xiong, Wenwen Qu, Peng Sun, Penglong Jiao, Han Lv, Lijun Wu, Kaipeng Zhang, Huipeng Deng, Jiaye Ge, Kai Chen, Limin Wang, Min Dou, Lewei Lu, Xizhou Zhu, Tong Lu, Dahua Lin, Yu Qiao, Jifeng Dai, and Wenhai Wang.
\newblock Internvl3: Exploring advanced training and test-time recipes for open-source multimodal models, 2025.

\end{thebibliography}
}

\clearpage

\appendix 
\setcounter{figure}{0} 
\setcounter{table}{0}

\section*{Overview of Appendix}

\begin{itemize}
    \item  \ref{appendix:Different}: \textbf{Different and Other Grounding-QA}
    \item  \ref{appendix:Holistic and KVs Detail}: \textbf{Holistic Video Representation Detail} 

    \item  \ref{appendix:ToTG-bench detail}: \textbf{ToTG-bench detail}
    
    \item  \ref{appendix:benchmark_details}: \textbf{Metric}
    \item  \ref{appendix:Experiment}: \textbf{Experimental Settings \& Additional Results}
    \item \ref{appendix:videoqadetail}: \textbf{VideoQA Task Details}
    \item \ref{appendix:Limitations & Future Work}: \textbf{Limitations \& Future Work}
    \item \ref{appendix:Broader Impacts Statement}: \textbf{Broader Impacts Statement}
    \item \ref{appendix:Qualitative Results}: \textbf{Qualitative Results}

\end{itemize}

\begin{figure*}[h]
    \centering
    \includegraphics[width=1.0\textwidth]{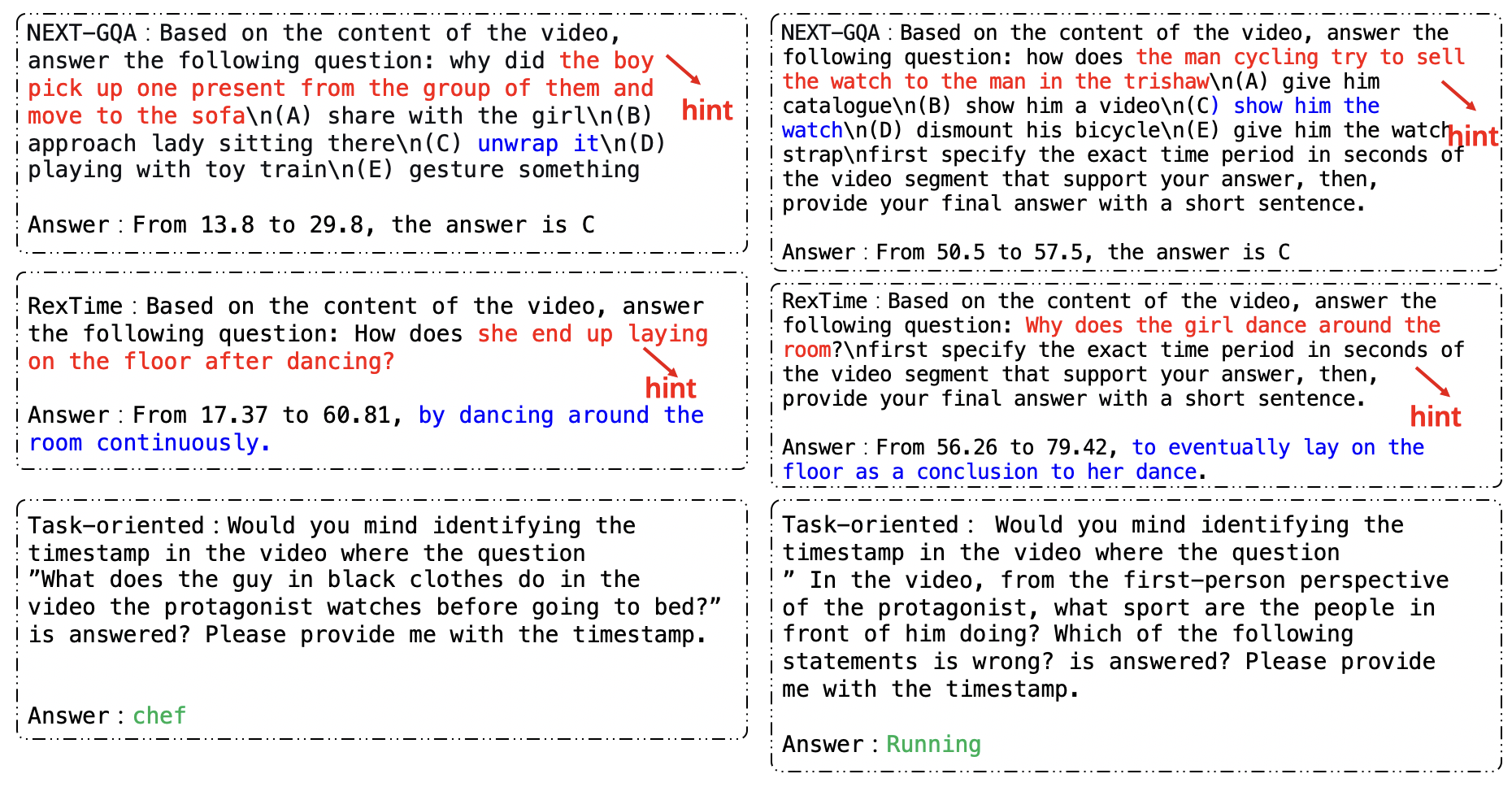}
    \caption{Different and Other Grounding-QA. Red denotes the hint to the answer span in the question, whereas blue marks the corresponding part in the answer. Task-oriented questions contain no cue about the target span, while all other grounding-QA tasks do.}
    \label{fig:totgdif}
    \vspace{-0.3cm}
\end{figure*}

\section{Different and Other Grounding-QA}
\label{appendix:Different}
Although some traditional groundingQA datasets—such as CG-Bench, NExT-GQA, and ReXTime already require localizing temporal evidence based on the question rather than on explicit event phrases, the Task-oriented Temporal Grounding task we propose is fundamentally different. These differences are illustrated in Figure~\ref{fig:totgdif}. Specifically, questions in conventional question-based grounding works usually still contain words strongly correlated with the target time interval, allowing existing temporal-grounding models to localize the interval by relying on the dominant cues in the question. In contrast, our task-oriented setting emphasizes reasoning and thinking driven by the question: our questions typically lack descriptors that are directly tied to the answer, and instead demand that the model first infer which temporal segments can solve the question and then ground them.



\section{Holistic Video Representation Detail}
\label{appendix:Holistic and KVs Detail}
We conducted a comprehensive ablation on the parameters of Holistic Video Representation (HVR), including the chunk size used to partition the video and the compression ratio applied when generating HVR. We obtain HVR from Detail Video Representation through a pooling operation. During training, we randomly select one-third of the samples to be trained with HVR. As shown in Tables~\ref{tab:holistic1} and ~\ref{tab:holistic2}, although using HVR leads to a slight performance drop when the video is input at 1 fps with a maximum of 800 frames, we observe clear performance gains when the maximum number of input frames is increased to 2,000. With HVR, the model surpasses the original zeroshot results.

\begin{table}[h]
\centering
\fontsize{8}{10}\selectfont
\setlength{\tabcolsep}{1.5pt}
\resizebox{0.5\textwidth}{!}{
\begin{tabular}{lcc|cc|cc}
\toprule 
\multicolumn{2}{c}{\textbf{Method}} &
\multirow{2}{*}{\textbf{Max Frames}} & 
\multicolumn{2}{c|}{\textbf{V-STaR(long)}} & \multicolumn{2}{c}{\textbf{ToTG-Bench(long)}} \\
VideoChunk & HVR Rate & & R1@0.5 & R1@0.7 & R1@0.5 & R1@0.7\\
\midrule
\rowcolor{gray!20}
0 & 1* & 800 & 73.2 & 61.4 & 37.8 & 29.5 \\
0 & 2*  & 800 & 63.6 & 63.6 & 31.2 & 27.7 \\
0 & 4* & 800 & 62.5 & 60.2 & 30.4 & 26.4 \\
150 & 1* & 800 & 61.3 & 60.2 & 29.5 & 26.1 \\
150  & 2* & 800 & 63.6 & 63.6 & 31.8 & 27.3 \\
150 & 4* & 800 & 62.5 & 61.3 & 30.6 & 26.9 \\
300 & 1* & 800 & 60.2 & 58.0 & 28.7 & 24.9 \\
300 & 2* & 800 & 63.6 & 63.6 & 32.4 & 26.3 \\
300 & 4* & 800 & 61.3 & 59.1 & 28.6 & 25.8 \\
\midrule
\bottomrule
\end{tabular}
}
\caption{The effectiveness of Holistic Video Representation.'0' indicates no video partitioning; '-' denotes CUDA out of memory (OOM).}
\label{tab:holistic1}
\end{table}

\begin{table}[h]
\centering
\fontsize{8}{10}\selectfont
\setlength{\tabcolsep}{1.5pt}
\resizebox{0.5\textwidth}{!}{
\begin{tabular}{lcc|cc|cc}
\toprule 
\multicolumn{2}{c}{\textbf{Method}} &
\multirow{2}{*}{\textbf{Max Frames}} & 
\multicolumn{2}{c|}{\textbf{V-STaR(long)}} & \multicolumn{2}{c}{\textbf{ToTG-Bench(long)}} \\
VideoChunk & HVR Rate & & R1@0.5 & R1@0.7 & R1@0.5 & R1@0.7\\
\midrule
\rowcolor{gray!20}
0 & 1* & 800 & 73.2 & 61.4 & 37.8 & 29.5 \\
0 & 1* & 2000 & - & - & - & - \\
150 & 1* & 2000 & 62.5 & 60.2 & 31.4 & 26.7 \\
150 & 2*  & 2000 & 80.7 & 76.1 & 41.0 & 37.1 \\
150 & 4* & 2000 & \textbf{86.4} & \textbf{81.8} & 43.2 & 39.7 \\
300 & 1* & 2000 & 64.7 & 62.5 & 33.4 & 28.2 \\
300 & 2* & 2000 & 79.5 & 78.4 & 40.6 & 37.3 \\
300 & 4* & 2000 & 81.8 & 80.6 & 42.7 & 38.8 \\
\midrule
\bottomrule
\end{tabular}
}
\caption{The effectiveness of Holistic Video Representation.'0' indicates no video partitioning; '-' denotes CUDA out of memory (OOM).}
\label{tab:holistic2}
\end{table}

\section{ToTG-bench detail}
\label{appendix:ToTG-bench detail}
\begin{figure*}[h]
    \centering
    \includegraphics[width=\textwidth]{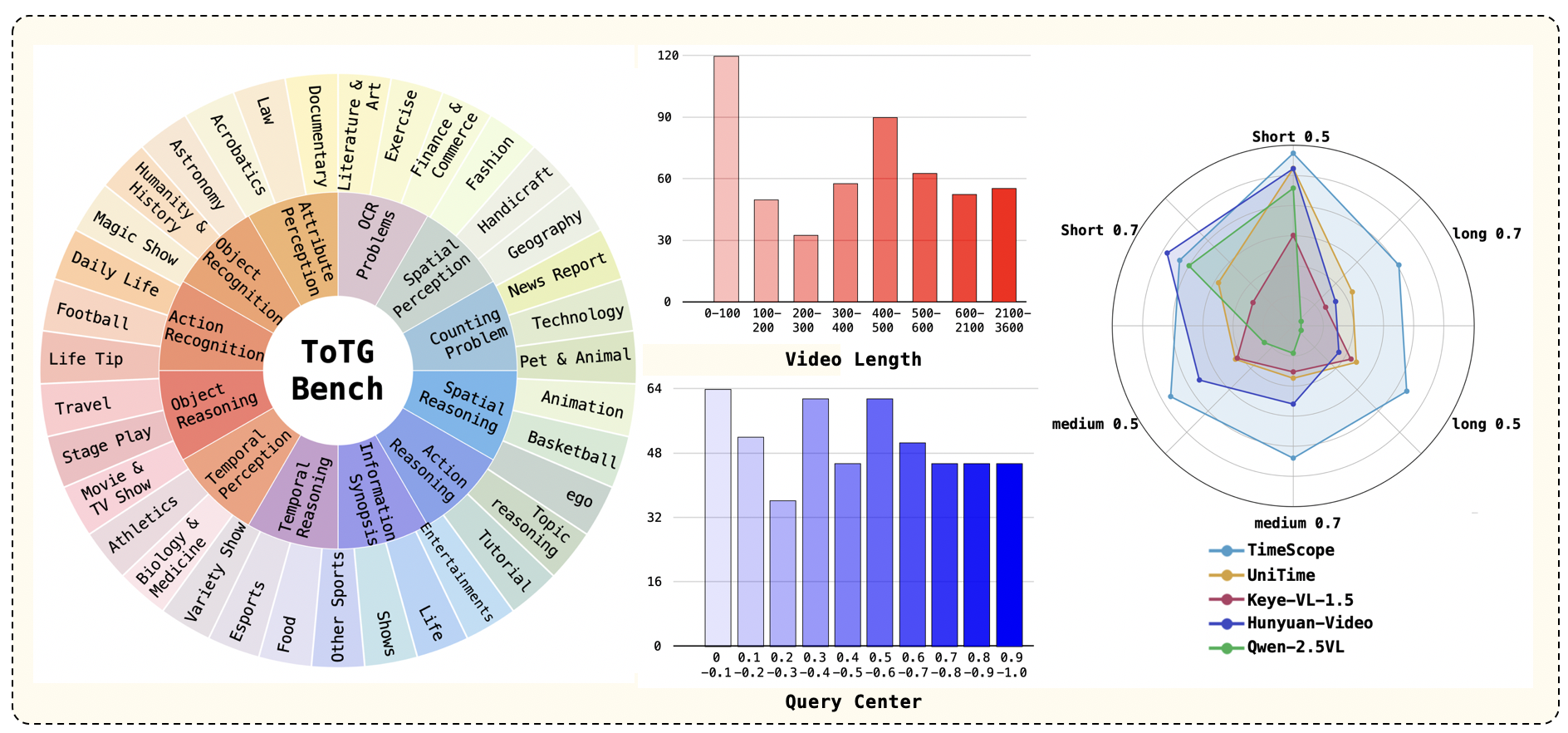}
    \caption{Statistics analysis of ToTG-bench. (Left) Our benchmark covers distinct task types and
35 video categories. (Middle) Video duration and question center distributions. (Right) Performance
of various model on ToTG-bench.}
    \label{fig:lvtg}
    \vspace{-0.3cm}
\end{figure*}

ToTG-Bench comprises 337 videos and nearly 500 questions. As shown in Figure~\ref{fig:lvtg}, it exhibits rich diversity in task types, video categories, video durations, and the temporal locations of target intervals. The average video length is 805 seconds. Specifically, the benchmark covers 13 task types—such as action reasoning, OCR perception, and temporal reasoning—spanning a wide range of video-understanding tasks.\footnote{We adopt the definitions of task type and video category from Video-MME.} In addition, it includes 35 video categories from numerous real-world domains. Video durations range from a few minutes to one hour, and the target intervals are uniformly distributed along the timeline to ensure an unbiased evaluation.

\textbf{Explanations of the 13 Task Categories in ToTG-Bench:}

\begin{itemize}
\item \textbf{OR (Object Reasoning)}: Identify why an object is present and how it causally affects later events in the clip.
\item \textbf{SR (Spatial Reasoning):} Infer spatial relationships (e.g., behind, left-of) between objects that are never simultaneously visible.

\item \textbf{TR (Temporal Reasoning)}: Determine the correct chronological order of events shown out of sequence or with missing segments.

\item \textbf{AR (Action Reasoning)}: Predict the next action an agent will perform given the current context and goals.

\item \textbf{SP (Spatial Perception)}: Locate and describe where objects are in the 3-D scene from egocentric or exocentric views.

\item \textbf{AP (Attribute Perception)}: Recognize object properties such as color, material, or state that remain constant across frames.

\item \textbf{TP (Temporal Perception)}: Detect when a change or event occurs and report its exact start and end times.

\item \textbf{EG (Ego)}: Understand what the camera-wearer is doing, attending to, or will do next from first-person video.

\item \textbf{OP (OCR Problems)}: Read and transcribe text that appears on screens, signs, or documents within the video.

\item \textbf{AT (Action Recognition)}: Classify which predefined action is being performed in a given segment of the video.

\item \textbf{CP (Counting Problem)}: Accurately count how many instances of a specified object or person appear throughout the clip.

\item \textbf{TU (Tutorial)}: Follow instructional video content and answer questions about the steps or techniques demonstrated.
\end{itemize}

\section{Metric}
\label{appendix:benchmark_details}

To evaluate the performance of our model on temporal grounding and VideoQA tasks, we employ the Intersection-over-Union (IoU) metric and its variants. These metrics quantify the alignment between the predicted temporal window \( T^{\text{pred}} \) and the ground truth temporal window \( T^{\text{gt}} \). The formal definitions are as follows:

\subsection{IoU and mIoU}
The fundamental IoU metric is defined as:
\[
\text{IoU} = \frac{|T^{\text{pred}} \cap T^{\text{gt}}|}{|T^{\text{pred}} \cup T^{\text{gt}}|}
\]

The mean Intersection-over-Union (mIoU) is calculated as the average IoU across all test samples:
\[
\text{mIoU} = \frac{1}{N} \sum_{i=1}^{N} \text{IoU}_i
\]
where \( N \) is the total number of test samples and \( \text{IoU}_i \) is the IoU value for the \( i \)-th sample.

We also evaluate performance using IoU thresholds, which measure the percentage of predictions exceeding specific IoU values:

\begin{table*}[h]
\centering
\setlength{\tabcolsep}{2pt}
\begin{tabular}{l|c|cccccccccc|c}
\midrule
\multicolumn{2}{c|}{\textbf{Query Center}} & \textbf{0-0.1} & \textbf{0.1-0.2} & \textbf{0.2-0.3} & \textbf{0.3-0.4} & \textbf{0.4-0.5} & \textbf{0.5-0.6} & \textbf{0.6-0.7} & \textbf{0.7-0.8} & \textbf{0.8-0.9} & \textbf{0.9-1} & diff\\ 
\midrule
\multirow{3}{*}{Qwen2.5VL} & R@0.3 & 0.417 & 0.368 & 0.615 & 0.222 & 0.250 & 0.158 & 0.278 & 0.235 & 0.133 & 0.154 & 78\%\\
                           & R@0.5 & 0.250 & 0.158 & 0.462 & 0.185 & 0.062 & 0.158 & 0.111 & 0.011 & 0.066 & 0.000 & 97\%\\
                           & R@0.7 & 0.208 & 0.158 & 0.308 & 0.074 & 0.000 & 0.053 & 0.055 &  0.000 & 0.000 & 0.000 & 100\%\\
\midrule
\multirow{3}{*}{Keye-1.5-VL} & R@0.3 & 0.615 & 0.652 & 0.666 & 0.444 & 0.375 & 0.353 & 0.556 & 0.600 & 0.235 & 0.461 & 64\%\\
                           & R@0.5 & 0.538 & 0.521 & 0.400 & 0.379 & 0.186 & 0.294 & 0.444 & 0.466 & 0.176 & 0.307 & 67.2\%\\
                           & R@0.7 & 0.462 & 0.260 & 0.266 & 0.259 & 0.125 & 0.176 & 0.222 &  0.333 & 0.117 & 0.231 & 74.6\%\\
\midrule
\multirow{3}{*}{Unitime} & R@0.3 & 0.669 & 0.465 & 0.633 & 0.364 & 0.371 & 0.216 & 0.550 & 0.567 & 0.371 & 0.567 & 67.7\%\\
                           & R@0.5 & 0.592 & 0.335 & 0.633 & 0.221 & 0.294 & 0.216 & 0.500 & 0.566 & 0.311 & 0.433 & 65.0\%\\
                           & R@0.7 & 0.476 & 0.335 & 0.367 & 0.186 & 0.176 & 0.163 & 0.500 &  0.455 & 0.194 & 0.433 & 67.4\%\\
\midrule
\multirow{3}{*}{Timescope} & R@0.3 & 0.393 & 0.417 & 0.606 & 0.464 & 0.421 & 0.470 & 0.555 & 0.667 & 0.470 & 0.467 & \textbf{35.1\%}\\
                           & R@0.5 & 0.393 & 0.417 & 0.606 & 0.357 & 0.368 & 0.353 & 0.505 & 0.556 & 0.412 & 0.400 & \textbf{41.7\%}\\
                           & R@0.7 & 0.357 & 0.375 & 0.533 & 0.286 & 0.263 & 0.294 & 0.400&  0.500 & 0.353 & 0.267 & \textbf{50.6\%}\\

\midrule
\end{tabular}
\label{tab:h}
\end{table*}

\begin{table}[h]
\centering
\vspace{-0.1in}
\renewcommand{\arraystretch}{1.15}
\begin{tabular}{>{\kern-0.5\tabcolsep}l|c|c<{\kern-0.5\tabcolsep}}
    \toprule
    \textbf{Hyperparameter} & \textbf{Stage 1} & \textbf{Stage 2} \\
    \midrule
    Overall batch size  & 64 & 64 \\
    Learning rate  & 1e-5 & 1e-5 \\
    LR Scheduler  & Cosine decay & Cosine decay \\
    DeepSpeed ZeRO Stage  & ZeRO-2-offload & ZeRO-2-offload \\
    Optimizer  & Adam & Adam \\
    Warmup ratio  & 0.3 & 0.3 \\
    Epoch  & 1 & 1 \\
    Weight decay  & 0 & 0 \\
    Precision  & bf16 & bf16 \\ 
    \bottomrule
\end{tabular}
\caption{Hyperparameters of Timescope for different training stages}
\label{tab:hyper}
\end{table}

\begin{table}[h]
\centering
    \centering
    \vspace{-0.1in}
    \renewcommand{\arraystretch}{1.15}

    \begin{tabular}{>{\kern-0.5\tabcolsep}l|c<{\kern-0.5\tabcolsep}}
        \toprule
        \textbf{Dataset}  & \textbf{Context Length}  \\
        \midrule
        Charades-STA  & 1fps \\
        Activity-Net  & 1fps \\
        Vid-Chapters  & 1fps($<$800 frames) \\
        CG-Bench & 1fps($<$800 frames) \\
        MLVU & 1fps($<$800 frames) \\
        LongVideoBench & 1fps($<$800 frames) \\
        V-STaR & 1fps($<$800 frames) \\
        \bottomrule
    \end{tabular}
    \caption{Experimental settings of Timescope.}
    \label{tab:benchmark}
\end{table}

\begin{itemize}
    \item \textbf{IoU@0.3}: Percentage of predictions with IoU $\geq$ 0.3
    \[
    \text{IoU@0.3} = \frac{1}{N} \sum_{i=1}^{N} \mathbb{I}(\text{IoU}_i \geq 0.3) \times 100\%
    \]
    
    \item \textbf{IoU@0.5}: Percentage of predictions with IoU $\geq$ 0.5
    \[
    \text{IoU@0.5} = \frac{1}{N} \sum_{i=1}^{N} \mathbb{I}(\text{IoU}_i \geq 0.5) \times 100\%
    \]
    
    \item \textbf{IoU@0.7}: Percentage of predictions with IoU $\geq$ 0.7
    \[
    \text{IoU@0.7} = \frac{1}{N} \sum_{i=1}^{N} \mathbb{I}(\text{IoU}_i \geq 0.7) \times 100\%
    \]
\end{itemize}

Here, \( |T^{\text{pred}} \cap T^{\text{gt}}| \) denotes the duration of the overlapping region between the predicted window \( T^{\text{pred}} \) and the ground truth window \( T^{\text{gt}} \). \( |T^{\text{pred}} \cup T^{\text{gt}}| \) represents the duration of their union, while \( |T^{\text{pred}}| \) and \( |T^{\text{gt}}| \) denote the durations of the predicted and ground truth windows, respectively. The indicator function \( \mathbb{I}(\cdot) \) equals 1 when the condition is true and 0 otherwise.

IoU measures the overall alignment between \( T^{\text{pred}} \) and \( T^{\text{gt}} \), providing a balanced assessment of both precision and recall by considering the overlap relative to their union. The threshold-based metrics (IoU@0.3, IoU@0.5, IoU@0.7) evaluate the model's ability to produce high-quality predictions meeting different precision standards, while mIoU provides an overall average performance measure across all samples.



\subsection{Query Center Robustness}
To measure the effectiveness of balanced query centers in the ToTG-benchmark, we evaluated the sensitivity of state-of-the-art (SOTA) models to query centers. As shown in the figure, experiments indicate that most models perform better on test data with query centers positioned earlier rather than later or in the middle. We measured the difference between the best and worst performance of various models when the query center varies. The results show that Timescope maintains its effectiveness regardless of the position of the query center (for instance, the gap is only 35\% at iou@0.3, which is significantly better than Qwen2.5vl's 78\% and 23.8\% higher than UniTime). This further proves the effectiveness of Timescope's training.

\section{Experimental Settings \& Additional Results}
\label{appendix:Experiment}
We elaborate on the training and inference details of Timescope. The reported hyperparameters cover Stage 1, and 2, as specified in Table~\ref{tab:hyper}. For the inference details, we emphasize the particular context length for different benchmarks, as shown in Table~\ref{tab:benchmark}.


\section{VideoQA Task Details}
\label{appendix:videoqadetail}

\subsection{VideoQA with Temporal Grounding}

We use \texttt{Qwen2-VL-7B} as the VideoQA model for answer generation. By default, it processes long videos by uniformly sampling 32 frames. However, this sampling strategy may lead to the omission of critical information. To investigate whether temporal grounding models can compensate for this issue, we adopt the following procedure. First, we use different video temporal grounding models to localize the relevant segments for each question. Then, we crop the localized video intervals and input them into \texttt{Qwen2-VL-7B}. Specifically, for cropped video segments shorter than 32 seconds, we extend their duration from the center to 32 seconds. Within each interval, we again uniformly sample 32 frames for answer generation.

\subsection{Prompt template for VideoQA}

We use the same prompt template for all multiple-choice VideoQA benchmarks:

\begin{verbatim}
System:
You are a helpful assistant.
User:
<video>
Question: <question>
Options:
(A) <Option_A>
(B) <Option_B>
(C) <Option_C>
(D) <Option_D>

Please only give the best option.
Best Option:
Assistant:
\end{verbatim}

\section{Limitations \& Future Work}
\label{appendix:Limitations & Future Work}

Although Timescope demonstrates exceptional performance on various video temporal grounding and video QA benchmarks, it still has several limitations that warrant further exploration: (i) Timescope is currently constrained to temporal grounding tasks (including traditional temporal grounding tasks and Task-Oriented Temporal Grounding tasks). To enable broader applications in MLLMs, it requires more diverse training data with dense temporal annotations. Incorporating such data into the pretraining process of MLLMs could unlock their potential for handling more temporally complex tasks, such as dense video captioning. (ii) Although Timescope enhances MLLMs with temporal grounding capabilities, relying solely on temporal grounding data limits their reasoning and question-answering abilities. The ultimate objective is to develop MLLMs that seamlessly integrate localization, reasoning, and question-answering into a unified framework.

\section{Broader Impacts Statement}
\label{appendix:Broader Impacts Statement}
Our research introduces a new problem, called Task-oriented Temporal Grounding (ToTG), to formally conceptualize the aforementioned challenge. We have created ToTG-Bench, aimed at evaluating temporal grounding performance in diverse real-world long video understanding scenarios and accelerating progress in this emerging field. This facilitates advancing the complexity of temporal understanding in videos and accelerates the development of models that integrate thinking and traditional temporal grounding capabilities. We hope that ultimately, the two can be integrated, unifying problem thinking and temporal grounding. We have also developed a more accurate and efficient temporal grounding framework, Timescope, to advance the field of long video temporal understanding. This could benefit a wide range of downstream applications, such as anomaly detection, security monitoring, etc. We believe that ToTG and Timescope can advance the development of video temporal understanding.

\section{Qualitative Results}
\label{appendix:Qualitative Results}
\begin{figure*}[h]
    \centering
    \includegraphics[width=\textwidth]{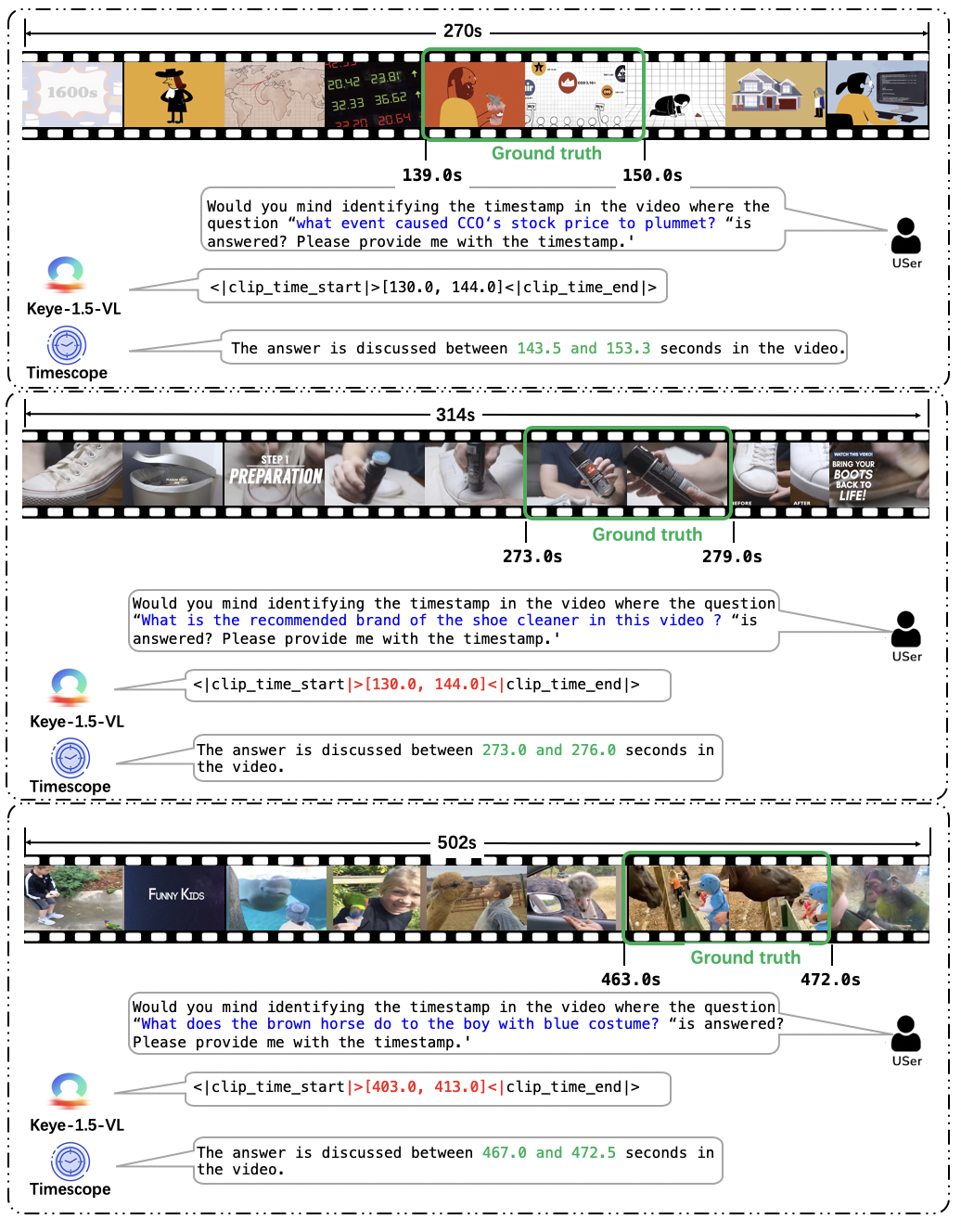}
    \caption{Qualitative Results of Timescope.}
    \label{fig:qr}
    \vspace{-0.3cm}
\end{figure*}

\begin{figure*}[h]
    \centering
    \includegraphics[width=1.1\textwidth]{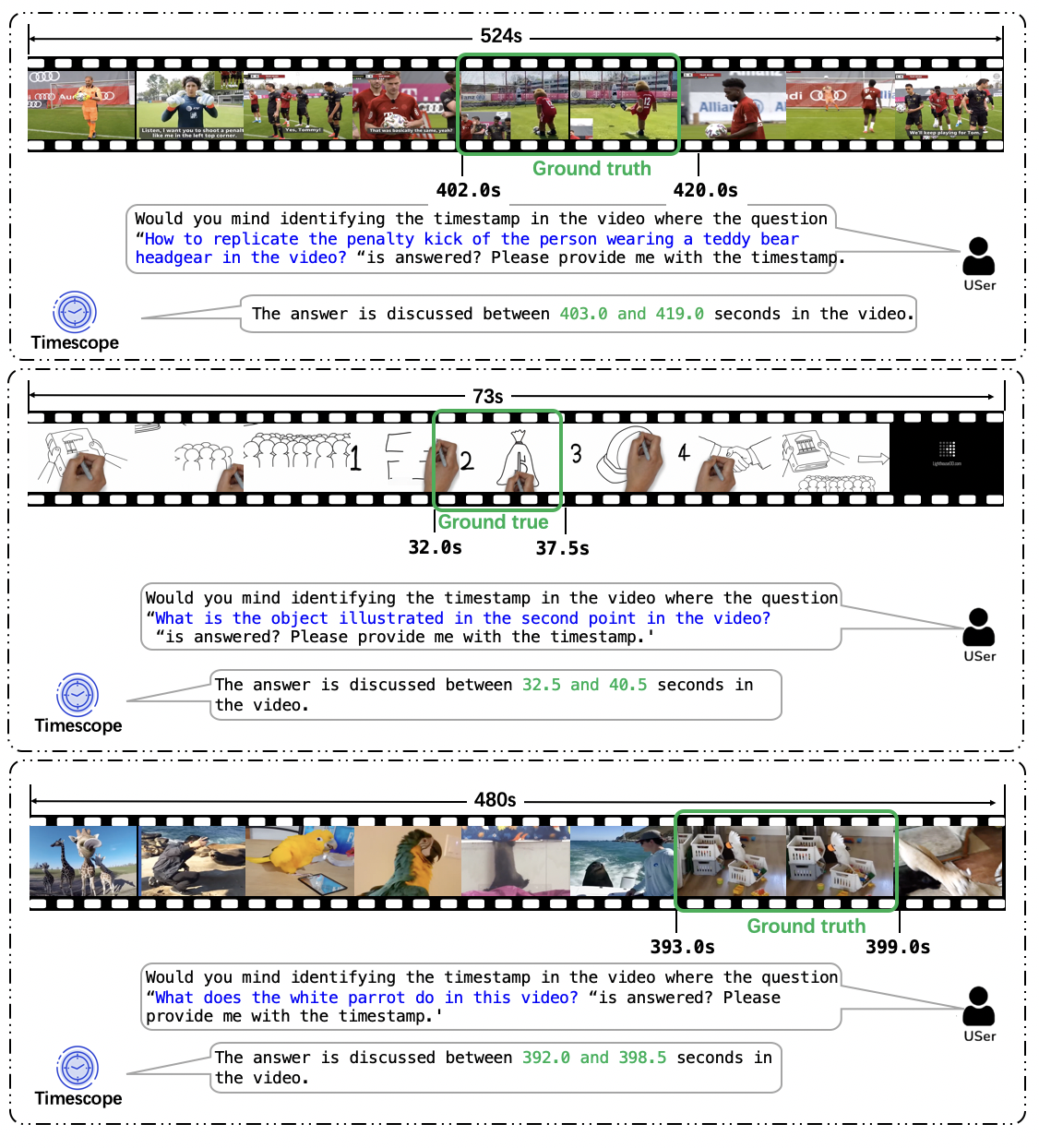}
    \caption{Qualitative Results of Timescope.}
    \label{fig:result1}
    \vspace{-0.3cm}
\end{figure*}

\begin{figure*}[h]
    \centering
    \includegraphics[width=1.1\textwidth]{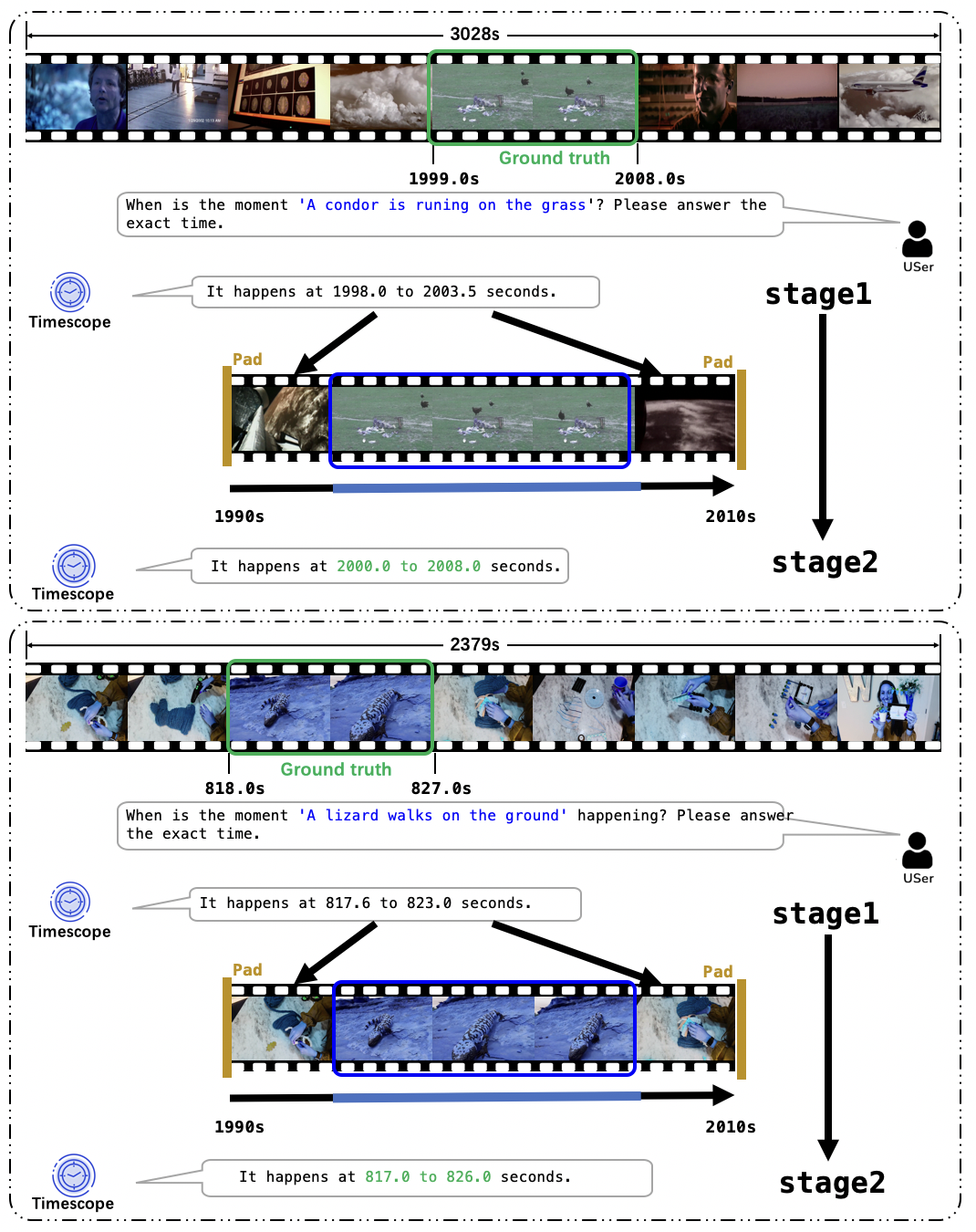}
    \caption{Qualitative Results of Timescope.}
    \label{fig:result2}
    \vspace{-0.3cm}
\end{figure*}

In Figure~\ref{fig:qr}--\ref{fig:result2}, we present the qualitative results of Timescope on ToTG-bench and V-STaR. Timescope demonstrates accurate understanding of questions and the ability to provide temporal grounding. In V-STaR, we show the results when two-stage reasoning is applied, and it can be seen that Timescope exhibits robust performance with good coarse-grained segment retrieval and fine-grained temporal grounding capabilities.

\begin{figure*}[h]
    \centering
    \includegraphics[width=\textwidth]{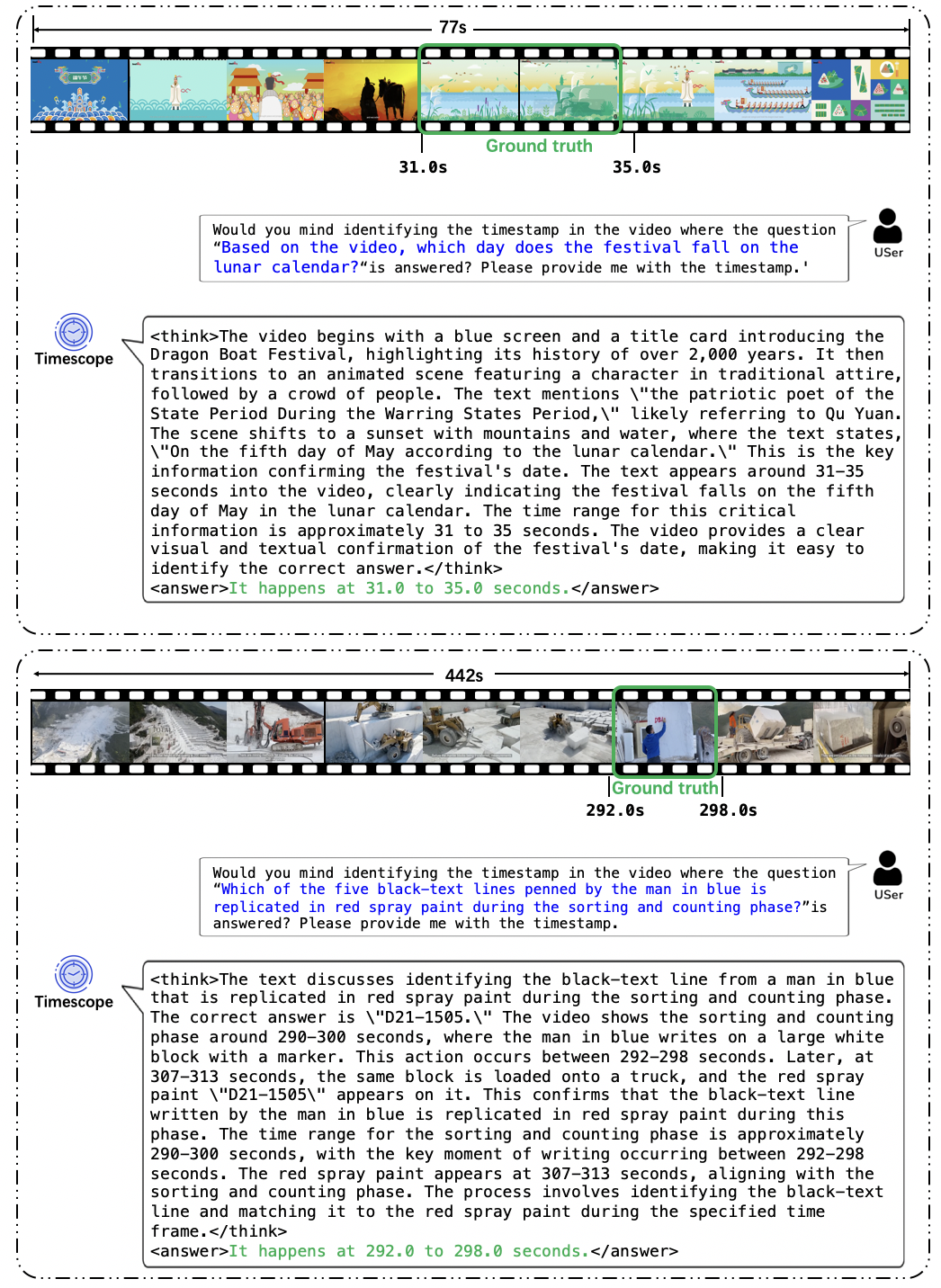}
    \caption{Qualitative Results of Timescope.}
    \label{fig:qr1}
\end{figure*}

\begin{figure*}[h]
    \centering
    \includegraphics[width=\textwidth]{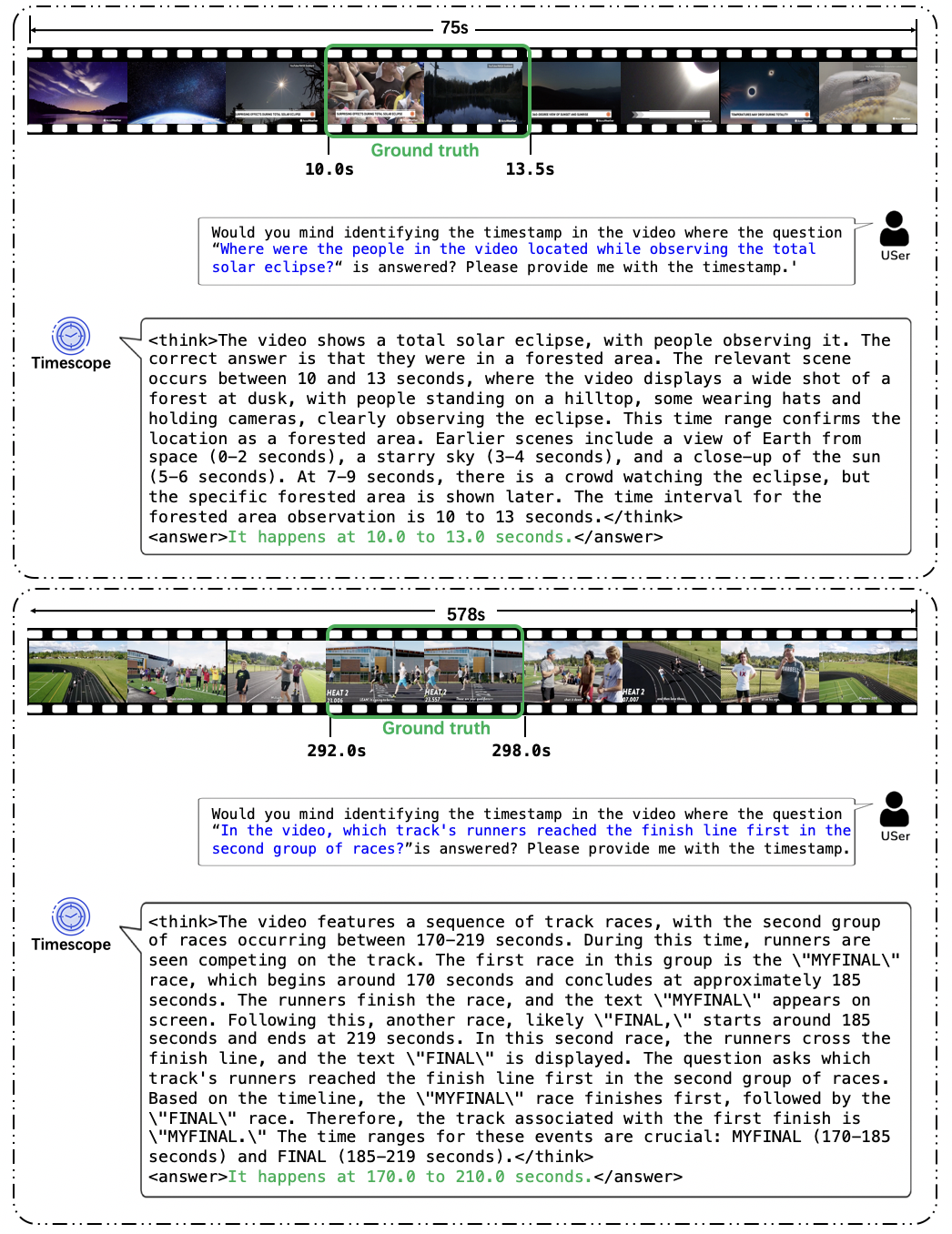}
    \caption{Qualitative Results of Timescope.}
    \label{fig:qr2}
\end{figure*}

\end{document}